\documentclass[sigconf,nonacm]{acmart}
\renewcommand\footnotetextcopyrightpermission[1]{}
\usepackage{url}
\usepackage{amsmath, amsfonts}
\usepackage[table]{xcolor}
 \usepackage{wrapfig}
\definecolor{bestgreen}{HTML}{E7FBFF}
\definecolor{secondbluebase}{HTML}{E7FBFF}
\definecolor{jclightblue}{HTML}{E5E5FA}
\colorlet{secondblue}{secondbluebase!40!white}

\usepackage{microtype}
\usepackage{graphicx}
\usepackage{multirow}
\usepackage{subcaption}
\usepackage{caption}
\usepackage{booktabs}
\usepackage{svg}
\usepackage{amsmath}
\usepackage{mathtools}
\usepackage{amsthm}
\usepackage{algorithm}
\usepackage{algpseudocode}

\AtBeginDocument{%
  }
\begin{document}

\title{MAR-GRPO: Stabilized GRPO for AR-diffusion \\ Hybrid Image Generation}

\author{Xiaoxiao Ma}
\affiliation{%
  \institution{University of Science and Technology of China}
  \city{Hefei}
  \country{China}}

\author{Jiachen Lei}
\affiliation{%
  \institution{AMAP, Alibaba Group}
  \city{Hangzhou}
  \country{China}}

\author{Tianfei Ren}
\affiliation{%
  \institution{University of Science and Technology of China}
  \city{Hefei}
  \country{China}}

\author{Jie Huang}
\affiliation{%
  \institution{University of Science and Technology of China}
  \city{Hefei}
  \country{China}}

\author{Siming Fu}
\affiliation{%
  \institution{Zhejiang University}
  \city{Hangzhou}
  \country{China}}

\author{Aiming Hao}
\affiliation{%
  \institution{AMAP, Alibaba Group}
  \city{Hangzhou}
  \country{China}}

\author{Jiahong Wu}
\affiliation{%
  \institution{AMAP, Alibaba Group}
  \city{Hangzhou}
  \country{China}}

\author{Xiangxiang Chu}
\affiliation{%
  \institution{AMAP, Alibaba Group}
  \city{Hangzhou}
  \country{China}}

\author{Feng Zhao}
\authornote{Corresponding author.}
\affiliation{%
  \institution{University of Science and Technology of China}
  \city{Hefei}
  \country{China}}

\renewcommand{\shortauthors}{Ma et al.}

\begin{abstract}
Reinforcement learning (RL) has been successfully applied to autoregressive (AR) and diffusion models. However, extending RL to hybrid AR--diffusion frameworks remains challenging due to interleaved inference and noisy log-probability estimation.
In this work, we study masked autoregressive models (MAR) and show that the diffusion head plays a critical role in training dynamics, often introducing noisy gradients that lead to instability and early performance saturation.
To address this issue, we propose a stabilized RL framework for MAR. We introduce multi-trajectory expectation (MTE), which estimates the optimization direction by averaging over multiple diffusion trajectories, thereby reducing diffusion-induced gradient noise. To avoid over-smoothing, we further estimate token-wise uncertainty from multiple trajectories and apply multi-trajectory optimization only to the top-$k\%$ uncertain tokens. In addition, we introduce a consistency-aware token selection strategy that filters out AR tokens that are less aligned with the final generated content.
Extensive experiments across multiple benchmarks demonstrate that our method consistently improves visual quality, training stability, and spatial structure understanding over baseline GRPO and pre-RL models. Our code is available at \href{https://github.com/AMAP-ML/mar-grpo}{\textcolor{blue}{https://github.com/AMAP-ML/mar-grpo}}.
\end{abstract}
\begin{teaserfigure}
  \centering
  \includegraphics[width=1\linewidth]{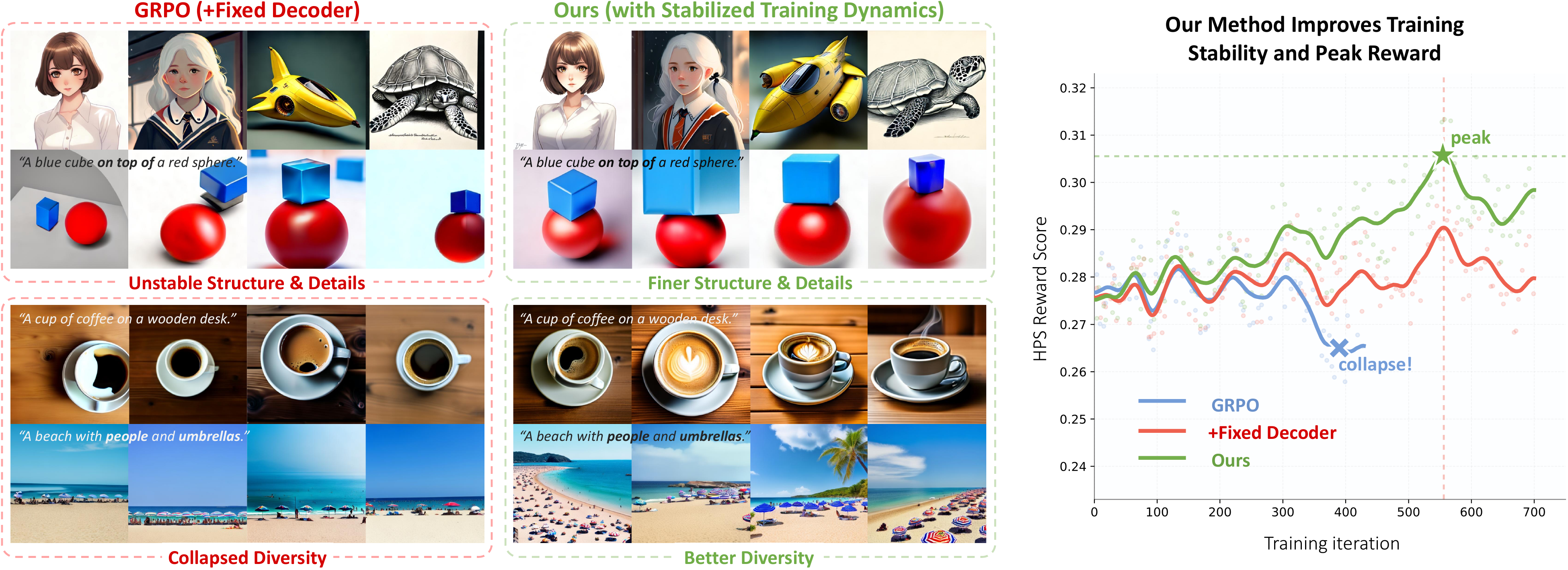}
    \caption{
    \textbf{Stabilizing MAR optimization via improved training dynamics.}
    \textbf{Left:} Compared to GRPO with a fixed decoder, standard optimization suffers from unstable structures, noisy details, and diversity collapse.
    \textbf{Middle:} Our method achieves stabilized training dynamics, producing more consistent structures, finer details, and improved diversity.
    \textbf{Right:} Training curves show that ours stabilizes reward dynamics and achieves higher peak performance, avoiding the collapse observed in baseline GRPO.
    }
  \label{fig_teaser}
\end{teaserfigure}
\maketitle
\section{Introduction}
Autoregressive models~\cite{chen2025janus_pro,sun2024llamagen,ma2025betterfasterautoregressive,zhang2025groupcriticaltokenpolicyoptimization} have recently attracted increasing attention in text-conditioned image synthesis due to their potential to unify text and image modalities~\cite{wu2024janus,xie2024show_o,deng2025bagel,xu2025scalar}. Among them, MAR has demonstrated strong performance by combining AR generation with continuous latent representations, mitigating the quantization bottleneck introduced by discrete tokenizers. 
In MAR, the AR transformer predicts continuous latent features in a \textit{masked autoregressive manner}, while a \textit{lightweight diffusion head} is applied at each AR step to refine latents and decode them into high-quality images. 
This hybrid design alleviates the limitations of discrete tokenization and has been widely explored in image~\cite{fan2024fluid}, video~\cite{deng2024autoregressive,yu2025videomar,teng2025magi}, and unified generation~\cite{team2025nextstep,wu2025harmonizing,kou2024orthus}.

Despite these advantages, MAR models still struggle with fine-grained prompt fidelity and visual quality, especially in complex or preference-sensitive scenarios. A well-established approach to address these issues is post-training with reinforcement learning (RL), where models are optimized using preference-based or verifiable rewards. 
In particular, Group Relative Policy Optimization (GRPO)~\cite{shao2024deepseekmath} has shown strong potential in improving both language models~\cite{deepseekai2025deepseekr1,yu2025dapo,chu2025gpg} and visual generative models~\cite{wang2025simplear,liu2025flow_grpo,jiang2025t2i_r1,sun2026varrlrighttackling,ma2025stagestablegeneralizablegrpo} by enabling stable and sample-efficient policy optimization.

Nevertheless, to the best of our knowledge, online reinforcement learning for the MAR-style hybrid paradigms remains largely unexplored and poses fundamental optimization challenges. First, MAR-style generation involves a complex inference process with multiple rounds of alternating AR and diffusion steps, leading to highly coupled trajectories and non-trivial log-probability estimation.
Second, the mismatch in parameter scale and complementary roles between large-scale AR transformer and lightweight diffusion head leads to significant instability in RL optimization.
As shown in Fig.~\ref{fig_motivation_fixhead}, \textit{naively applying vanilla GRPO with end-to-end manner often leads to instable optimization and early performance saturation}, we attribute it to disparate optimization dynamics inherent in joint AR and diffusion frameworks. To further investigate this issue,  given the lightweight nature of diffusion head and its limited role in content generation, we analyze and compare the training dynamics of standard end-to-end GRPO with a variant that fixes diffusion head and optimizes only the AR transformer.

As a result, optimizing only the AR component yields more stable and smoother training dynamics than end-to-end optimization, consistently improving image quality and prompt-following capability. 
Interestingly, although the diffusion head primarily refines low-level details of latents produced by the AR model, it has a significant impact on overall training dynamics. 
We attribute this to the large gradient variance introduced by the stochastic denoising process, where the model must handle samples with different noise levels across multiple timesteps.
From this perspective, the diffusion decoder can be viewed as a conditional decoder that heavily depends on AR latents, while its inherent stochasticity introduces noise into the optimization process. 
Concretely, during GRPO rollout, given a fixed AR latent, sampling multiple diffusion trajectories and estimating the optimization direction via expectation over these trajectories can significantly improve training stability (See Fig.~\ref{fig_visualize_ab_seed}).

Based on the above analysis, we propose MAR-GRPO, the first GRPO-based method tailored for the AR--diffusion hybrid paradigm. 
Specifically, we introduce three key components. 
(1) We leverage multiple diffusion trajectories obtained during rollout and estimate the optimization direction via expectation over these trajectories, leading to more stable optimization dynamics. 
(2) Based on token-wise uncertainty from multiple diffusion samples, we apply multi-trajectory estimation only to the top-$k\%$ most uncertain tokens, achieving a better trade-off between stability and computational efficiency with negligible additional overhead due to the lightweight diffusion head. 
(3) A consistency-aware token selection strategy is applied to filter out tokens whose optimization directions are inconsistent with the final generated content, further improving visual detail and stability.
Our contributions are as follows:

\begin{figure*}[t]
\setlength{\abovecaptionskip}{0.1cm}
\setlength{\belowcaptionskip}{0.1cm}
\centering
\begin{minipage}[b]{.44\linewidth}
    \begin{minipage}[b]{.5\linewidth}
        \includegraphics[width=1.\linewidth]{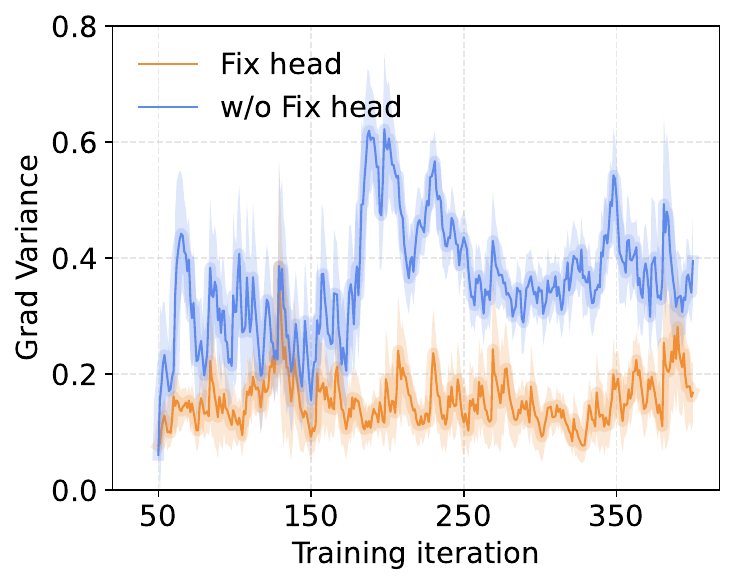}
    \end{minipage}\hfill
    \begin{minipage}[b]{.5\linewidth}
        \includegraphics[width=1.\linewidth]{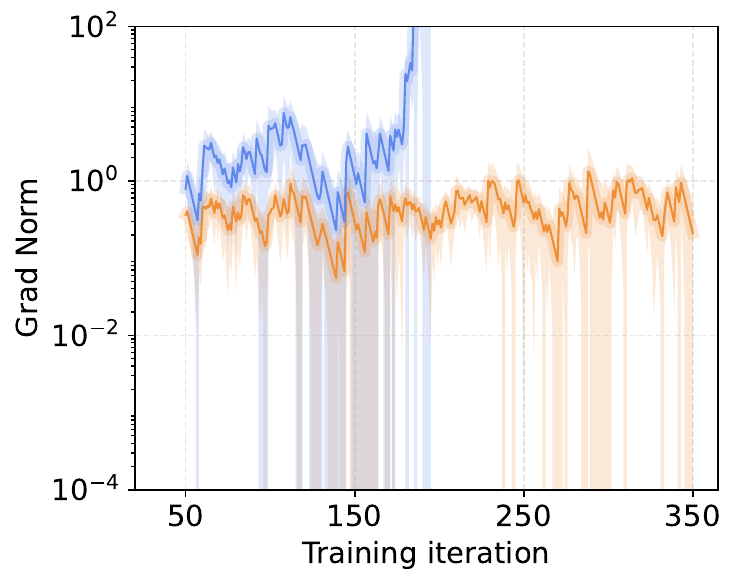}
    
    \end{minipage}
    \subcaption{Effect of freezing diff. head on grad. variance and norm.}
    \label{fig:grad_var_norm_fix_head}
\end{minipage}\hfill
\begin{minipage}[b]{.22\linewidth}
    \includegraphics[width=1.\linewidth]{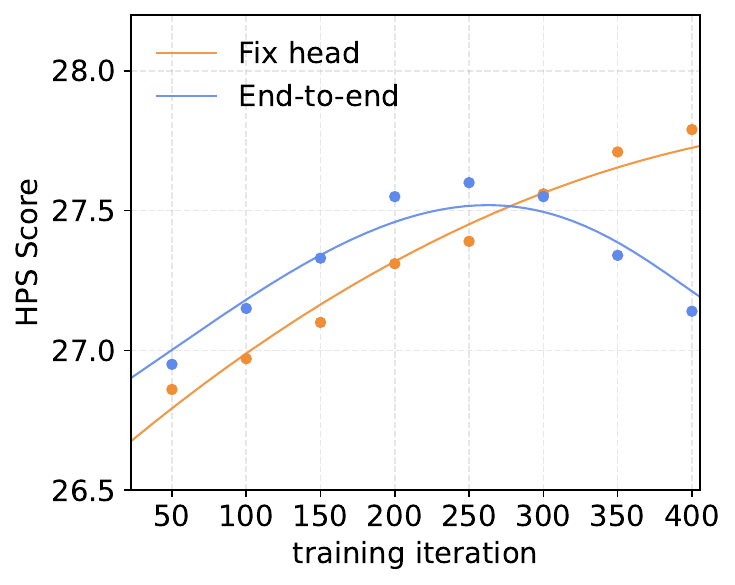}
    \subcaption{Benchmarking HPS score.}
    \label{fig:hps_head}
\end{minipage}\hfill
\begin{minipage}[b]{.31\linewidth}
    \includegraphics[width=1.\linewidth]{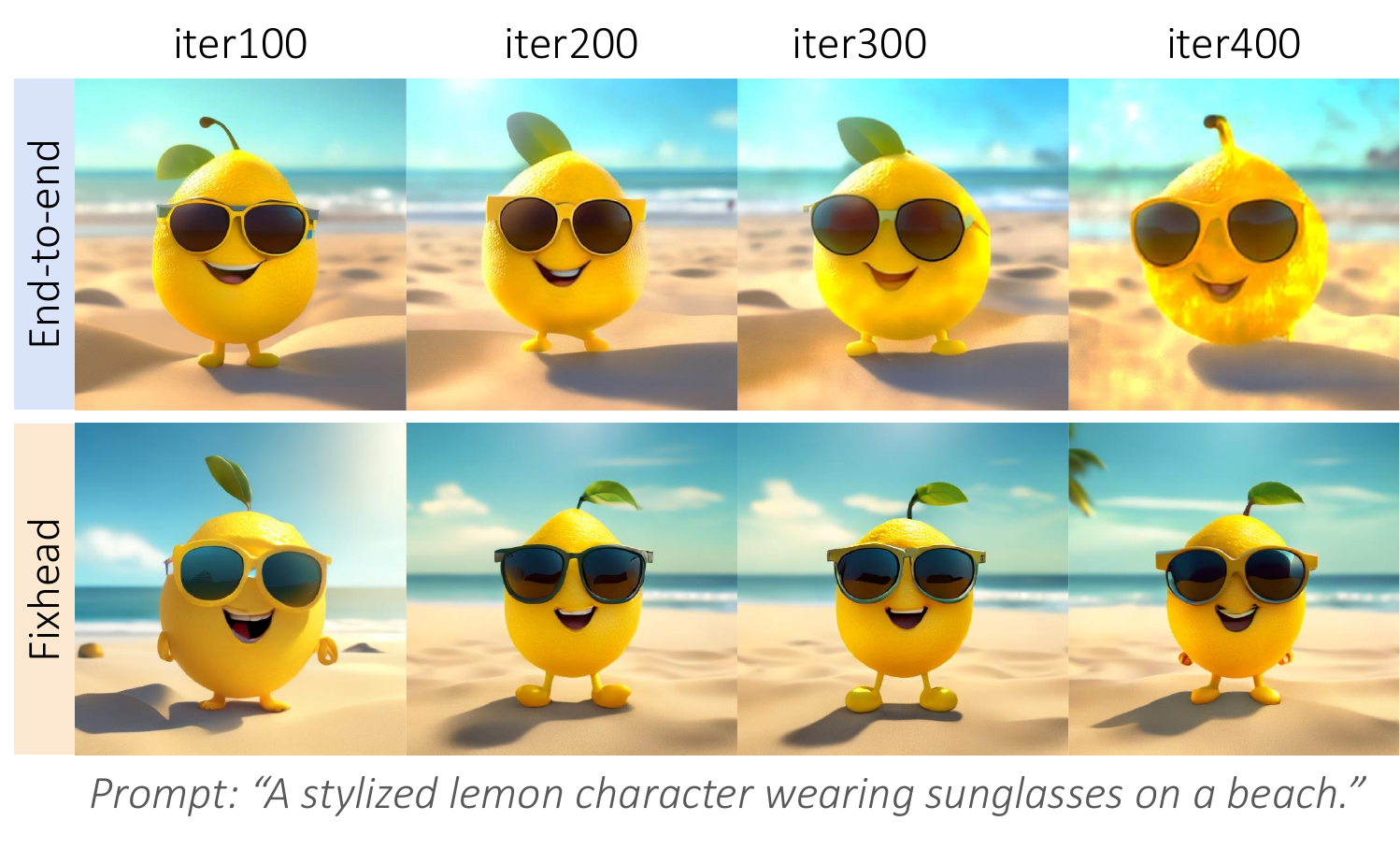}
    \subcaption{Visual impact of diffusion head fixing.}
\end{minipage}
 \caption{
    (a) Gradient comparison between the end-to-end GRPO baseline and the frozen diffusion head counterpart. The End-to-end baseline introduces "noisier" gradients to the AR component, disturbing the training progress.
    (b) Test set HPS v.s. training iteration. HPS curves show that end-to-end baseline suffers from reward hacking and early saturation.
    (c) End-to-end training degrades image quality in later stages, which can be addressed by freezing the diffusion head.
    } 
    \label{fig_motivation_fixhead}
\end{figure*}

\begin{enumerate}
    \item We present the first GRPO-based framework for masked autoregressive (MAR) models, and identify the diffusion head as a key source of optimization instability in hybrid AR--diffusion architectures.
    \item Based on analysis of the training dynamics under this framework, we propose multiple trajectory based optimization, which computes expectation of multiple diffusion trajectories, reducing diffusion-induced variance and yielding more stable optimization dynamics.
    \item Compared to baseline GRPO methods and pre-RL models, our approach achieves consistent improvements across benchmarks on human preference, prompt following, and spatial structure understanding.
\end{enumerate}

\section{Related Works}
\subsection{AR-diffusion Hybrid Paradigms}
Autoregressive (AR) models generate images by predicting tokens sequentially in raster order~\cite{chen2025janus_pro,liu2024lumina_mgpt} or mask-based manner~\cite{bai2024meissonic} over a Vector-Quantized (VQ)~\cite{esser2021vqgan} discrete space. Such approaches are often limited by the quantization accuracy of the tokenizer, making it difficult to synthesize high-fidelity images. Mask autoregressive (MAR)~\cite{li2024autoregressive} models address this issue by decoding AR latents with a lightweight model via flow matching~\cite{flux2024}, effectively avoiding discrete quantization errors, and their potential has been explored in prior studies. Fluid~\cite{fan2024fluid} first extends this paradigm to text-conditioned image generation. FAR~\cite{yu2025frequencyautoregressiveimagegeneration} explores combining scale-wise autoregressive paradigm~\cite{tian2024var,ma2024star} with diffusion head, while subsequent works such as Harmon~\cite{wu2025harmonizing} further investigate hybrid generation for unified understanding and synthesis. NOVA~\cite{deng2024autoregressive} and VideoMAR~\cite{yu2025videomar} extend MAR to video generation, and~\cite{zou2025fastardiff} explores acceleration. Notably, NextStep~\cite{team2025nextstep} scales the AR backbone to a $14$B-parameter transformer, demonstrating the strong scalability of AR-diffusion hybrid paradigm. However, online reinforcement learning for MAR-based generation remains unexplored.

\subsection{RL for Visual Generation}
Motivated by recent advances in Reinforcement Learning with Verifiable Reward (RLVR) for large language models (LLMs)~\cite{deepseekai2025deepseekr1,shao2024deepseekmath,yu2025dapo,chu2025gpg}, RL-based methods, especially Group Relative Policy Optimizatio (GRPO) have been increasingly studied for visual generation.
Specifically, for continuous generation with diffusion models, Flow-GRPO~\cite{liu2025flow_grpo} and Dance-GRPO~\cite{xue2025dance_grpo} analyze the challenges of applying GRPO to flow-based models, while subsequent works~\cite{he2025tempflow,li2026mixgrpo} explore time-step-based reward assignment strategies to further improve training efficiency. For discrete AR models, SimpleAR~\cite{wang2025simplear} and AR-GRPO~\cite{yuan2025ar_grpo} provide basic RL baselines, while T2I-R1~\cite{jiang2025t2i_r1} introduces an RL framework augmented with textual chain-of-thought (CoT). Follow-up studies further investigate token-level advantage assignment based on the semantic information encoded in tokens~\cite{zhang2025gcpo,ma2025stagestablegeneralizablegrpo} for improving efficiency. In addition, works such as Mask-GRPO~\cite{luo2025maskgrpo} and MaskFocus~\cite{zhang2025maskfocusfocusingpolicyoptimization} explore GRPO for mask-based AR generation, and NextFlow~\cite{sun2026varrlrighttackling} proposes a scale-wise AR generation strategy under the RL framework.
\section{Methods}


\begin{figure*}[t]
\setlength{\abovecaptionskip}{0.1cm}
\setlength{\belowcaptionskip}{0.1cm}
\centering
\includegraphics[width=\linewidth]{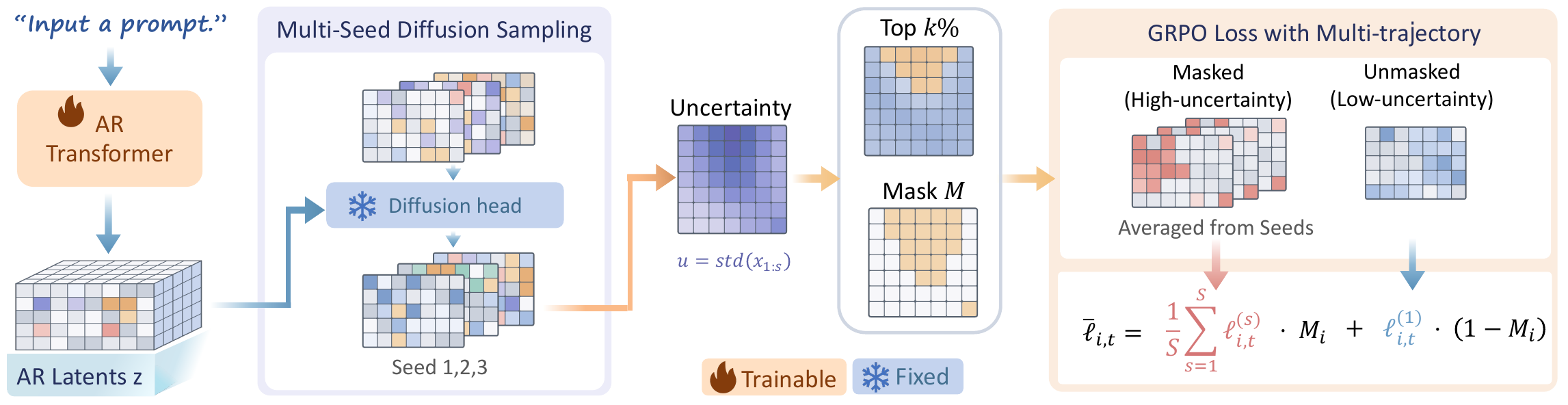}
    \caption{
    The proposed multi-trajectory expectation estimates an uncertainty map by sampling multiple diffusion trajectories during rollout, capturing regions that are most sensitive to diffusion-induced stochasticity. 
    Based on a top-$k\%$ uncertainty threshold, our noise-corrected optimization selectively aggregates multi-trajectory signals to calibrate the GRPO loss, effectively reducing stochastic noise in gradient estimation and yielding a more reliable optimization direction.
    }
    \label{fig_framework}
\end{figure*}

\subsection{Preliminaries}
\noindent\textbf{Masked autoregressive models.}
\label{sec_preliminary_ardiff}
Given a sequence of tokens $\{x^1, x^2, ..., x^n\}$ where the superscript specifies specific order, it formulates the generation problem as \textit{multiple next token prediction} over $K$ mask steps:
\begin{equation}
     p(X^1, ..., X^K) = \prod_{k=1}^K p(X^k | X^1, ..., X^{k-1}),
\end{equation}
where $X^k=\{x^i, x^{i+1}, ..., x^j\}$ is the set of latent tokens generated at the $k$-th mask step, with $\cup_k x^k=\{x^1, ..., x^n\}$.
MAR contains two model components: an AR model and a diffusion model, which is a light-weight MLP head. The AR model produces a set of conditioning vectors $Z^k$ at the $k$-th mask step by operating on previous tokens: $Z^k=f(X^1, ..., X^{k-1})$.
Subsequently, conditioned on $Z^k$, the diffusion model performs $T$ denoising steps to generate the final latent tokens $X^k$. We denote the intermediate diffusion latent at denoising step $t$ by $X_t^k$, where $X_0^k=X^k$. The AR and diffusion head are trained end to end via simple denoising loss.
Let $x_t=\sqrt{\alpha} x_{t-1} + \sqrt{1-\alpha}\epsilon$, $\epsilon\sim\mathcal{N}(0, \mathbf{I})$, $\alpha\in R$ and $t\in[0, T]$, $x_0\sim p_{data}(x)$, $x_T\sim \mathcal{N}(0, \mathbf{I})$ denote the forward process. The denoising loss can be written as:
\begin{equation}
\mathcal{L}_{MAR} = \mathbb{E}_{\epsilon, t}\biggl[\Vert \epsilon - \epsilon_\theta(x_t \mid t, z)\Vert^2\biggr].
\label{eq:mar_loss}
\end{equation}
We omit the superscript of AR output token $z$ for simplicity.


The sampling process typically proceeds through alternating rounds between AR model and diffusion head. Specifically, the diffusion head generates refined tokens conditioned on current AR outputs, after which the AR model predicts the next set of tokens based on those refined by diffusion head. 

\noindent\textbf{Group Relative Policy Optimization}.
GRPO updates image generation models through an iterative \emph{generate–evaluate–update} procedure.
In each optimization step, given prompt $c$, the policy  produces a group of images $\{o_1, ..., o_G\}$, where we retain all intermediate sampling states. 
Subsequently, the policy $\pi_\theta$ is optimized by maximizing the following objective:
\begin{equation}
\resizebox{1.0\linewidth}{!}{$
\begin{aligned}
\mathcal{J}_{\text{GRPO}}(\theta)
= \underset{\substack{c, i\in[1, G], \\ k\in[1, K]}}{{\huge\mathbb{E}}}\Bigg[
\min\!\Big(
    r_{i,k}(\theta)\hat{A}_{i}, \text{clip}_\epsilon(r_{i,k}(\theta)) \hat{A}_{i}
\Big)
 - \beta\, D_{\text{KL}}\!\big(\pi_\theta \,\|\, \pi_{\text{ref}}\big)
\Bigg],
\end{aligned}
$}
\label{eq_grpo_opt_objective}
\end{equation}
where $r$ is the importance ratio, $\text{clip}_\epsilon(r_{i,k}(\theta))=\text{clip}\!\big(r_{i,k}(\theta), 1-\varepsilon,\,1+\varepsilon\big)$, $\hat{A}_i$, commonly known as the advantage, is a scalar value assigned to each sample.
In our case, the importance ratio $r$ is defined at each AR mask step $k$ over the corresponding diffusion-generated latent tokens $X_i^k$ of sample $o_i$. Specifically, the ratio is defined as:
\begin{equation}
    r_{i,k}(\theta) = \tfrac{\pi_\theta(X_{i}^{k}\mid X_{i}^{1:k-1};c)}{\pi_{\theta_\text{old}}(X_{i}^{k}\mid X_{i}^{1:k-1};c)}.
    \label{eq_important_ratio}
\end{equation}
Here, $\pi_\theta$ is the current policy and $\pi_{\theta_{\text{old}}}$ is the sampling policy from the previous GRPO optimization step.
Since MAR generates each token group $X^k$ through an iterative diffusion process, the policy likelihood is further decomposed into the full denoising trajectory. Given the tokens $X^{1:k}$ derived from history steps, the probability of generated $X^k$ is:
\begin{equation}
\pi(X^k \mid X^{1:k-1}; c) \approx \prod_{t=1}^{T} p_\theta(X_{t-1}^k \mid X_t^k, Z^k),
\label{eq:policy}
\end{equation}
where each transition $p_\theta(X_{t-1}^k \mid X_t^k, Z^k)$ is parameterized by the diffusion head. 
The advantage $\hat{A}_i$ is calculated on normalized rewards within each sample group:
\begin{equation}
\hat{A}_{i} = \frac{R_i - \text{mean}(\{R_i\}_{i=1}^G)}{\text{std}(\{R_i\}_{i=1}^G)}.
\label{eq_preliminary_advantage}
\end{equation}
Here, $R_i=\mathcal{R}(o_i, c)$ is reward (e.g, HPS or CLIP score) of sample $o_i$ assigned by the reward model $\mathcal{R}$, which acts as a metric for evaluation the quality of images during rollout.



\subsection{Analysis \& Motivation}
\noindent\textbf{A simple RL baseline for optimizing MAR.}
Due to the absence of an existing GRPO baseline for MAR, we first describe a simple baseline approach to clarify how GRPO methods for diffusion models like Flow-GRPO~\cite{liu2025flow_grpo} or DanceGRPO~\cite{xue2025dance_grpo} can be adapted to optimize MAR models.
\label{sec_mar_grpo_baseline}

As described in Sec.~\ref{sec_preliminary_ardiff}, MAR generates samples by interleaving the sequential prediction of autoregressive models with the iterative denoising characteristic of diffusion models.
This interleaved, multi-step sampling process generates a rich set of intermediate activations, which are retained to facilitate subsequent policy gradient computations.
As a result, computing the GRPO over the entire sampling trajectory is computationally expensive.

As indicated in \eqref{eq:policy}, since the joint likelihood of the Markovian trajectory factorizes into independent transition probabilities, the global objective decomposes into a sum of step-wise losses. This allows us to optimize the model via stochastic approximation.
Specifically, we approximate $\pi(X^k \mid X^{1:k-1}; c)$ by randomly selecting a subset of AR mask steps and denoising timesteps along the entire sampling trajectory $\{X^1, ..., X^K\}$.
As a result, the importance ratio and KL regularization term are computed on the selected subsets. Let $\ell(\theta)=\min\!\big(r_{i,m}(\theta)\hat{A}_{i},\,\mathrm{clip}_\varepsilon(r_{i,m}(\theta))\hat{A}_{i}\big)$ denote the step-wise term in Eq.~\eqref{eq_grpo_opt_objective}, where $m$ indexes a selected AR mask step.
The objective $\mathcal{J}_\text{GRPO}$ becomes:
\begin{equation}
\mathcal{J}_\text{GRPO}(\theta) = \mathbb{E} \biggl[ \frac{1}{|\mathcal{M}| |\mathcal{T}|}\sum_{m \in\mathcal{M}}\sum_{t \in \mathcal{T}}\ell(\theta)\biggr].
\label{eq_baseline_mar}
\end{equation}

During training, following Sec.~\ref{sec_preliminary_ardiff}, for each selected AR step $m$, we compute the corresponding latent $z_m$ conditioned on $\{z_{m-1}, x_{t+1}\}$ collected from the rollout stage at diffusion timestep $t$, and evaluate log-probabilities over the selected diffusion timesteps. The loss is computed over all tokens predicted at the current step, rather than only those accepted at subsequent AR steps, which improves token utilization and training efficiency.

\noindent\textbf{Analysis on the baseline training dynamics.}
\label{sec_difference}
Optimizing MAR under RL introduces a coupled optimization between the AR model and the diffusion head.
Since AR latents do not admit explicit likelihoods, gradients must be propagated through the diffusion head along the denoising trajectory.
This induces a deep and stochastic gradient path, where diffusion noise is accumulated and backpropagated to the AR model.
Combined with the parameter scale mismatch between the large AR backbone and the lightweight diffusion head, such coupling can lead to amplified and unstable updates.

Specifically, based on GRPO baseline settings, we optimize the MAR model end-to-end and analyze the training dynamics. We reveal that diffusion head impact the training stability as indicated by a large gradient variance and norm.
As displayed in Fig.~\ref{fig:grad_var_norm_fix_head}, during training, the gradient of the baseline model fluctuates intensively, with its norm consistently increasing, showing the training is quite unstable. As a result, the human preference score (HPS) of the baseline model declines noticeably after 250 training iterations on HPSv2 benchmark prompts, see Fig.~\ref{fig:hps_head}.

We speculate two underlying factors contributing to the training instability:
(1) \textbf{Non-stationary reward signal.}
Due to the coupled optimization between the AR model and the diffusion head, the effective mapping from AR latents to final outputs continuously changes during training.
This makes the reward signal non-stationary, leading to ambiguous credit assignment for the AR model.
(2) \textbf{Denoising-induced gradient variance.}
The policy gradient is propagated through the diffusion head, which serves as a stochastic decoder from AR latents to image space.
Different noise realizations can lead to inconsistent outputs and rewards for the same latent, resulting in high-variance and potentially amplified gradients received by the AR model.
Below, we verify these hypotheses and address the instability with simple yet effective strategies with negligible computational overhead.

\subsection{Expectation of Diffusion Trajectories for Stabilized MAR Optimization}
\label{sec_mte}

\noindent\textbf{Diffusion head amplifies gradient noise.}
As discussed in Sec.~3.2, the coupled optimization between the AR model and the diffusion head introduces non-stationarity in the reward signal, making credit assignment difficult.
From an optimization perspective, updating the diffusion head continuously changes the mapping from AR latents to final outputs, resulting in a moving target for the AR model.
Consequently, the AR model cannot reliably attribute reward variations to its own token predictions, leading to unstable optimization dynamics.

Fixing the diffusion head effectively removes this source of non-stationarity by stabilizing the transformation from AR latents to image space.
This provides the AR model with a consistent optimization target and enables more reliable policy updates.
In addition, it eliminates gradient perturbations introduced by updates to the diffusion parameters, leaving only the inherent stochasticity of the denoising process. Empirically, we observe that freezing the diffusion head leads to significantly more stable training dynamics compared to end-to-end optimization.
As shown in Fig.~\ref{fig:grad_var_norm_fix_head} and Fig.~\ref{fig:hps_head}, the gradient statistics become smoother and the reward improves consistently over training.

We also explore reducing the learning rate of the diffusion head as a softer alternative.
However, even small updates to the diffusion head introduce noticeable disturbances to AR policy learning, suggesting that the optimization is highly sensitive to changes in the diffusion mapping.
Therefore, fixing the diffusion head provides a simple yet effective way to stabilize MAR optimization.

\noindent\textbf{Reducing gradient variance with multi-trajectory estimation.}
\label{sec_diffusion_key}
Even though we obtain a more stationary optimize signal by fixing the diffusion head in RL, the randomness of the diffusion head itself will still affect the optimization process.

From a reinforcement learning perspective, stable policy optimization relies on low-variance gradient estimates.
However, in MAR, the policy is implicitly defined through a diffusion-based decoder, where the likelihood is computed along stochastic denoising trajectories.
This induces a noisy gradient estimator, as the same AR latent can correspond to multiple possible decoded outputs under different noise realizations.
In this sense, the diffusion head defines a conditional distribution over outputs given AR latents, rather than a deterministic mapping.
Directly using a single trajectory therefore leads to a high-variance estimate of the policy likelihood.

\begin{figure}[t]
\setlength{\abovecaptionskip}{0.1cm}
\setlength{\belowcaptionskip}{0.1cm}
\centering
\begin{minipage}[b]{.48\linewidth}
    \includegraphics[width=\linewidth]{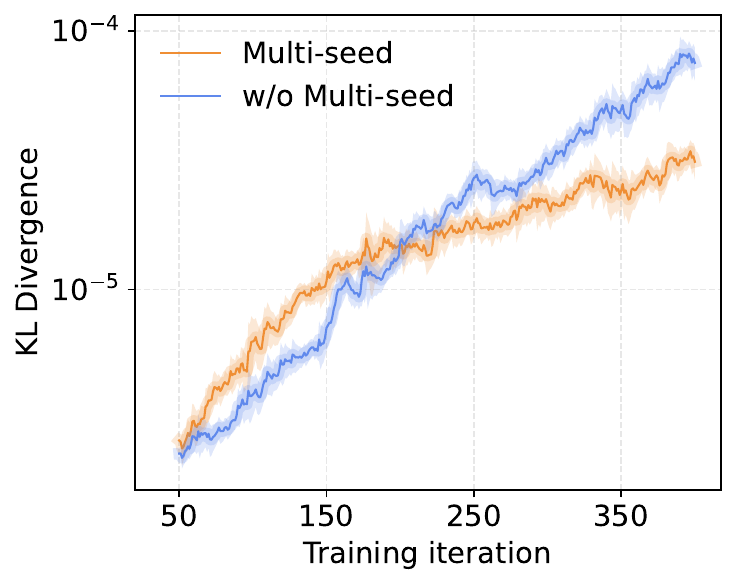}
\end{minipage}\hfill
\begin{minipage}[b]{.48\linewidth}
    \includegraphics[width=\linewidth]{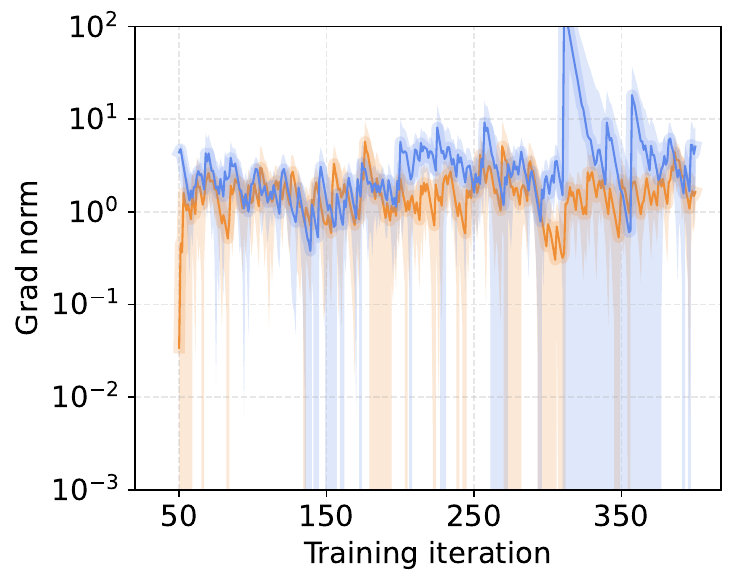}
\end{minipage}
\caption{
Compared to GRPO with fixed diffusion head, incorporating multi-seed trajectory expectation further corrects the optimization direction, leading to better
training dynamics (smoother KL loss growth and lower, more stable
gradient norms in later stages).
}
\vspace{-0.3cm}
\label{fig_visualize_ab_seed}
\end{figure}

Interestingly, we observe that diffusion trajectories conditioned on the same AR latents tend to produce highly similar latent structures, indicating that the semantic content is largely determined by the AR model, while the diffusion head primarily refines low-level details.
As a result, variations across diffusion trajectories mainly reflect stochastic decoding noise rather than meaningful differences in generation.
This suggests that averaging over multiple trajectories can effectively reduce estimation noise without significantly altering the underlying data distribution (See Fig.~\ref{fig_visualize_ab_seed}).

Specifically, we replace the single-trajectory likelihood with an expectation over multiple diffusion trajectories conditioned on the same AR latents:
\begin{equation}
    \hat{\pi}(X^{k}_i \mid X_i^{1:k-1}, c) \approx \mathbb{E}_{\epsilon} \Big[ \prod_{t=1}^T p_\theta(X^{k}_{i,t-1} \mid X^{k}_{i,t}, Z_i^k) \Big].
\end{equation}
We approximate this expectation using $S$ diffusion trajectories conditioned on the same $Z_i^k$.
This formulation can be interpreted as a variance-reduced estimator of the policy likelihood under diffusion stochasticity, leading to a more stable and reliable optimization direction.

\begin{figure}[t]
\setlength{\abovecaptionskip}{0.1cm}
\setlength{\belowcaptionskip}{0.1cm}
\begin{center}
\includegraphics[width=1\linewidth]{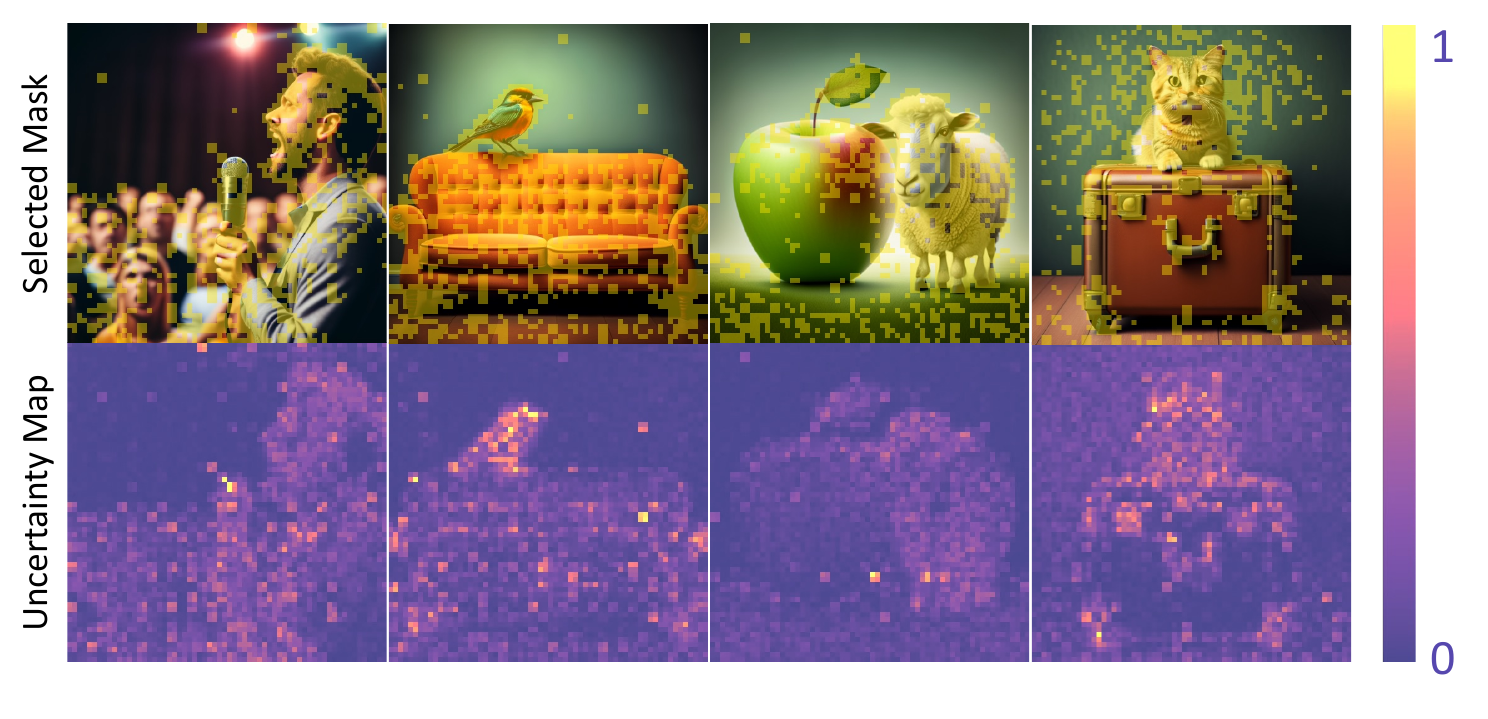}
\end{center}
\caption{
    Visualization of estimated uncertainty and corresponding uncertainty
    maps (bottom) with generated images (top). We highlight top
    $30\%$ most uncertain regions for clarity.
}
\vspace{-0.3cm}
\label{fig_critic_token}
\end{figure}

\noindent\textbf{Improving training efficiency via diffusion decoder uncertainty.}
While multi-trajectory estimation reduces gradient variance and stabilizes training, applying it uniformly to all tokens may be inefficient and can limit peak performance. 
Averaging over multiple trajectories may introduce a smoothing effect that weakens informative gradients.

This suggests that diffusion-induced uncertainty is not uniformly distributed: stochasticity tends to concentrate in regions with ambiguous semantics or fine-grained structures, while other regions remain relatively stable.

To capture this property, we estimate a token-wise uncertainty map using the variance across multiple diffusion trajectories:
\begin{equation}
u_i = \mathrm{std}\big(x_i^{1:S}\big),
\label{eq_token_std}
\end{equation}
where $x_i^{1:S}$ denotes the predictions at position $i$ from $S$ trajectories.
High values of $u_i$ indicate positions that are sensitive to diffusion stochasticity and thus prone to noisy gradient estimation.

Based on this observation, we apply multi-trajectory estimation selectively.
Specifically, we construct a binary mask $M_i$ by selecting the top-$k\%$ tokens with highest uncertainty from $u_i$, and apply expectation-based estimation only to these positions, while retaining the standard single-trajectory objective for the rest:
\begin{equation}
\mathcal{J}_{\text{GRPO}}(\theta)
=
\mathbb{E}\Big[
M_i \cdot \frac{1}{S}\sum_{s=1}^S \ell^{(s)}
+
(1 - M_i)\cdot \ell^{(1)}
-
\beta D_{\mathrm{KL}}
\Big].
\label{eq_margrpo}
\end{equation}

As shown in Fig.~\ref{fig_critic_token}, we visualize several cases of critic tokens.
High-uncertainty tokens are highlighted in yellow, and we observe that regions with rich structural information tend to exhibit higher variance.
This yields a selective variance reduction scheme that mitigates noise in uncertain regions while preserving sharp and informative gradients in stable regions.
As a result, our method improves training stability without sacrificing peak performance.

\begin{algorithm*}[t]
\caption{GRPO with proposed multi-trajectory expectation from diffusion decoder}
\label{alg_uncertainty_grpo}
\begin{algorithmic}[1]
\Require Selected AR mask step $k$, AR latent $Z_i^k$, sampled diffusion trajectories $\{X_{i,t}^{k,(s)}\}_{s=1}^{S}$, selected diffusion steps $\mathcal{T}$, number of trajectories $S$, uncertainty ratio $k\%$
\Ensure Uncertainty-aware GRPO loss for optimization

\State Sample $S$ diffusion trajectories conditioned on the same AR latent $Z_i^k$.
\For{selected AR mask steps $k \in \mathcal{M}$}
    \State Compute token-wise uncertainty $u_i^k$ following Eq.~\eqref{eq_token_std}.
    \State Construct binary mask $M_i^k$ by selecting the top-$k\%$ tokens according to $u_i^k$.
    \For{selected diffusion steps $t \in \mathcal{T}$}
        \State Compute step-wise losses $\ell_{i,k,t}^{(s)}$ using Eq.~\eqref{eq_important_ratio}, Eq.~\eqref{eq:policy}, and Eq.~\eqref{eq_preliminary_advantage}.
        \State Aggregate the uncertainty-aware loss according to Eq.~\eqref{eq_margrpo}.
    \EndFor
\EndFor
\State Update the policy based on Eq.~\eqref{eq_margrpo}.
\end{algorithmic}
\end{algorithm*}

\subsection{Consistency-Aware Token Selection}
\label{sec_consist_token_selection}

Beyond log-probability estimation in the diffusion head, the AR model also plays a critical role in RL optimization. In particular, among the tokens generated during rollout, not all tokens contribute positively to the final generation. Indiscriminately optimizing over all tokens may therefore introduce noise and degrade training efficiency.

To address this, we assess the consistency of AR predictions across masking steps and selectively include tokens that positively contribute to the final output.
For AR step $m$, let $x_0^m$ denote the predicted diffusion latent at step $m$, and $X_0^{-1}$ denote the final diffusion latent obtained from the complete AR latents.
For each spatial token (we omit index $j$ for simplicity), we define $\mathrm{sim}(m)$ as the cosine similarity between the corresponding tokens in $x_0^{m}$ and $X_0^{-1}$.

We use the change in similarity between consecutive AR steps as a criterion for token selection. Specifically, we construct a token-wise binary mask $\kappa^{(m)}$ for each AR step:
\begin{equation}
\kappa^{(m)} = \mathbb{I}\!\left[\Delta \mathrm{sim}(m) > \tau\right],
\end{equation}
where $\Delta \mathrm{sim}(m) = \mathrm{sim}(m) - \mathrm{sim}(m-1)$,
$\mathbb{I}[\cdot]$ denotes the indicator function, and $\tau$ is a predefined similarity threshold.
Here, $\kappa^{(m)} \in \{0,1\}$ indicates whether a token at step $m$ participates in policy optimization.
Accordingly, the loss term $\ell_{m,t}$ in Eq.~(\ref{eq_margrpo}) is weighted by $\kappa^{(m)}$, yielding $\kappa^{(m)} \ell_{m,t}$, so that only tokens with consistent improvements are emphasized during optimization.

\begin{figure*}[ht!]
\setlength{\abovecaptionskip}{0.1cm}
\setlength{\belowcaptionskip}{0.1cm}
\begin{center}
    \includegraphics[width=0.86\textwidth]{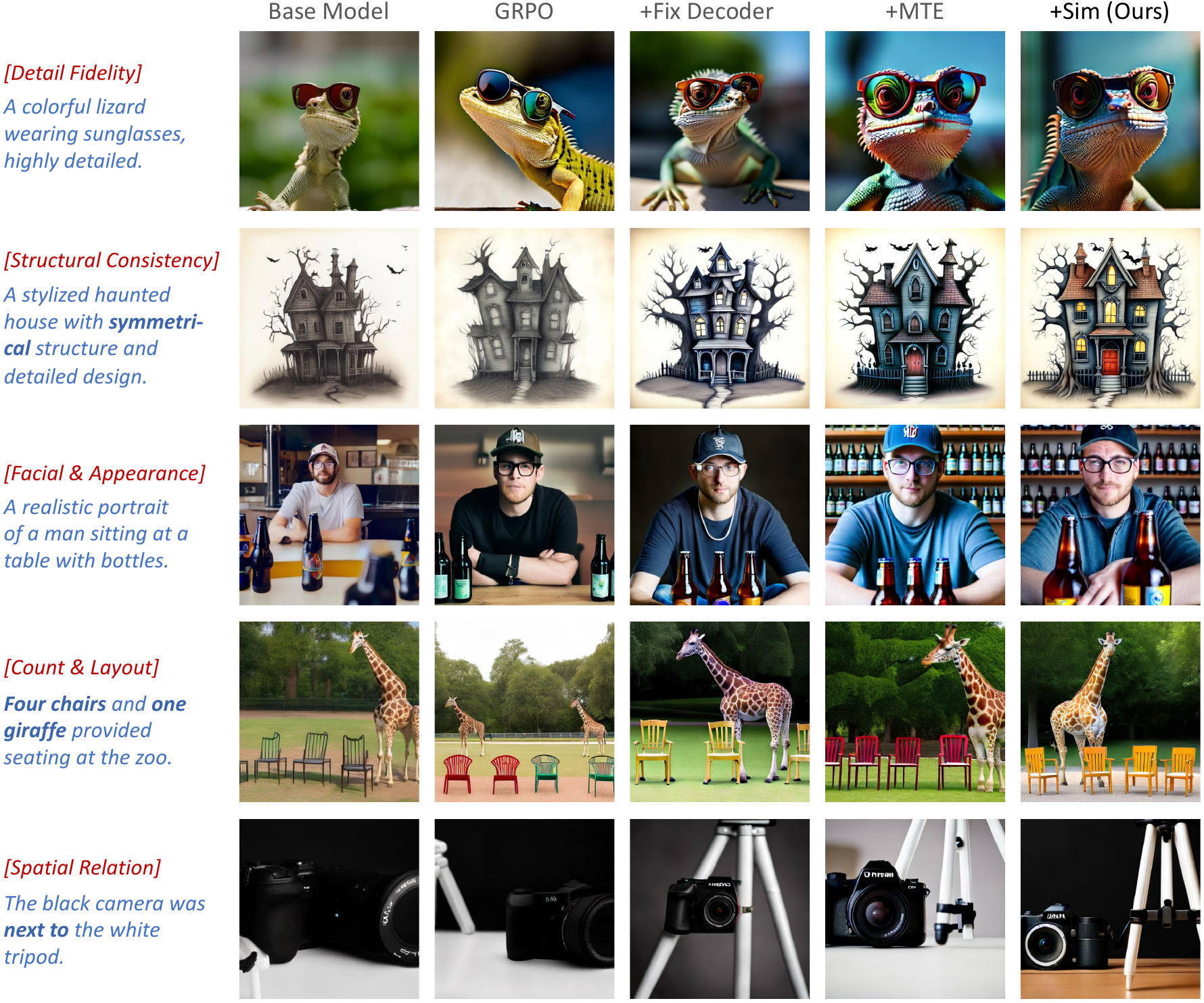}
\end{center}
\caption{
\textbf{Progressive improvement across multiple generation aspects.}
From left to right: base model, GRPO, +Fixed Decoder, +MTE, and our full model.
GRPO leads to unstable optimization, while fixing the decoder improves stability.
MTE further enhances visual quality by reducing diffusion-induced stochasticity.
Our full model, with consistency-aware token selection, achieves the best results in detail, structure, appearance, counting, and spatial relations.
}
\vspace{-0.2cm}
\label{fig_example_hps_spatial}
\end{figure*}

\section{Experiments}
\noindent\textbf{Training setup}. 
We choose NOVA~\cite{deng2024autoregressive} and Harmon~\cite{wu2025harmonizing} as our base models and adopt their default sampling settings in our experiments.
Specifically, for NOVA, we use a classifier-free guidance (CFG) scale of $5$, $64$ AR sampling steps, and $25$ diffusion sampling steps. We use the official $0.6$B model trained at $512\times512$ resolution.
For Harmon, we use CFG $=3$, $64$ AR sampling steps, and replace the original $100$-step diffusion sampler with a $25$-step DDIM sampler for efficiency, which has negligible impact on evaluation results. 
Following~\cite{zhang2025maskfocusfocusingpolicyoptimization}, we improve the AR sampling by identifying tokens that may hinder optimization via consistency-aware token selection (see Appendix for details). We ablate this design later.

For GRPO training, we build upon T2I-R1~\cite{jiang2025t2i_r1}, using a batch size of
$3$, group size $4$ and $\beta$=0.01 for KL regularization. To balance
training cost and performance, we randomly sample $12$ masking steps and $10$
diffusion steps from each rollout for policy optimization.
Additional experiment details are provided in the Appendix.

\noindent\textbf{Training Data \& reward}.
Following T2I-R1~\cite{jiang2025t2i_r1}, we train the model using short prompts that cover both single-object and multi-object scenarios, including compositional relationships such as object positions and counts. 
Such prompt design simplifies the generation space while emphasizing key aspects of visual reasoning, enabling more effective learning of spatial and numerical relationships. 
Moreover, it has been shown to improve aesthetic quality as well as spatial and numerical coherence in multi-object generation, making it particularly suitable for RL-based optimization.

We evaluate the models on preference-related metrics including
HPS~\cite{wu2023hpsv2} and ImageReward, as well as on T2I-CompBench~\cite{huang2023t2i_compbench},
which focuses on spatial and compositional reasoning. For preference metrics, we
report results from models trained with the HPS reward. For T2I-CompBench, we
follow T2I-R1 and adopt a mixed reward consisting of HPS, GIT, and GroundingDINO.

\begin{table*}[htbp]
\centering
\setlength{\abovecaptionskip}{0.1cm}
\setlength{\belowcaptionskip}{0.1cm}
\caption{Quantitative comparison results on the human preference correlated metrics (HPS, Imagereward, Pickscore and Aesthetic) and T2I-compbench. Our method ("Ours") achieves consistent gains across multiple metrics, significantly outperforming the standard GRPO implementation (“+GRPO”) as well as AR and diffusion baselines in visual quality and structural consistency. ``spat." is short for spatial. $^\dagger$: We observe failure cases in the base model, see Appendix for detailed analysis.}
\setlength{\tabcolsep}{2.5mm}{
\resizebox{\linewidth}{!}{
\begin{tabular}{lccccc|ccccccc}
\toprule
\multirow{2}{*}{Model} & \multirow{2}{*}{Param} & \multicolumn{4}{c|}{Preference} & \multicolumn{6}{c}{T2I-compbench} \\ \cmidrule(l){3-13} 
 &  & HPS & ImgRwd & Pickscore & Aesthetic & Color & Shape & Texture & 2d-spat. & 3d-spat. & Numeracy & Complex \\ \midrule
SD3 & 2B & 30.22 & 0.95 & 22.41 & 5.90 & 0.8094 & 0.5864 & 0.7297 & 0.3219 & 0.4044 & 0.6078 & 0.3780 \\
FLUX.1-dev & 12B & 31.35 & 1.10 & 22.77 & 6.27 & 0.7407 & 0.5718 & 0.6922 & 0.2863 & 0.3866 & 0.6185 & - \\
Sana-1.5 & 1.6B & 30.36 & 1.01 & 22.60 & 6.07 & 0.7625 & 0.5426 & 0.6761 & 0.3814 & 0.4088 & 0.6110 & 0.3727 \\ \midrule
LlamaGen & 0.8b & 23.92 & -0.38 & 20.50 & 5.21 & 0.2996 & 0.3212 & 0.3888 & 0.1004 & 0.1530 & 0.2747 & 0.2501 \\
Show-o & 1.3B & 27.98 & 0.86 & 22.00 & 5.90 & 0.7327 & 0.5264 & 0.6815 & 0.3697 & 0.3996 & 0.6209 & 0.3572 \\
Janus-Pro & 7B & 28.64 & 0.76 & 21.83 & 5.68 & 0.6355 & 0.3494 & 0.4929 & 0.1931 & 0.3279 & 0.4423 & 0.3566 \\ \midrule
NOVA & 0.6B & 26.76 & 0.44 & 21.57 & 5.79 & 0.6907 & 0.5489 & 0.6712 & 0.2983 & 0.3886 & 0.5933 & 0.3510 \\
~~+GRPO & - & 27.57 & 0.38 & 21.51 & 5.78 & 0.6721 & 0.5392 & 0.6690 & 0.3069 & 0.3984 & 0.5970 & 0.3484 \\
\rowcolor{jclightblue} ~~Ours & - & \textbf{29.35} & \textbf{0.64} & \textbf{21.69} & \textbf{6.01} & \textbf{0.7176} & \textbf{0.5780} & \textbf{0.6980} & \textbf{0.3306} & \textbf{0.4046} & \textbf{0.6027} & \textbf{0.3575} \\
Harmon$^\dagger$ & 1.5B & 25.76 & 0.15 & 21.35 & 5.66 & 0.6396 & 0.4104 & 0.5547 & 0.3153 & 0.3723 & 0.5551 & 0.2871  \\
~~+GRPO & - & 29.15 & 0.62 & 21.70 & 5.93 & 0.7408 & 0.4952 & 0.6135 & 0.3484 & 0.4140 & 0.6008 & 0.3208 \\
\rowcolor{jclightblue} ~~Ours & - & \textbf{29.57} & \textbf{0.65} & \textbf{21.80} & \textbf{5.95} & \textbf{0.7416} & \textbf{0.4993} & \textbf{0.6267} & \textbf{0.3547} & \textbf{0.4207} & \textbf{0.6096} & \textbf{0.3302} \\ \bottomrule
\end{tabular}
\label{tab_comp_perference}
}}
\vspace{-2mm}
\end{table*}

\subsection{Main Results among Metrics}
We evaluate the proposed method on human preference benchmarks, including HPS~\cite{wu2023hpsv2}, ImageReward~\cite{xu2024imagereward}, PickScore~\cite{Kirstain2023pickscore}, and Aesthetic Score, as well as on spatially correlated benchmarks such as T2I-CompBench~\cite{huang2023t2i_compbench}.

\noindent\textbf{Human preference.}
We evaluate models trained with HPS reward on preference-related metrics. As shown in Fig.~\ref{fig_example_hps_spatial} and Table~\ref{tab_comp_perference}, compared to the original pre-RL base model and baseline GRPO, the proposed method with fixing diffusion head, multiple diffusion trajectories and consistency token selection achieves significant improvements in both quantitative metrics and visual quality.

Moreover, compared to baseline GRPO with fixed diffusion decoder, our approach yields a more stable RL process (See Fig.~\ref{fig_teaser} and Fig.~\ref{fig_visualize_ab_seed}). Consequently, the post-RL model improves aesthetic quality while preserving image fidelity and structural consistency, thereby mitigating reward hacking to a certain extent.

\noindent\textbf{Spatial-correlated metrics.}
Beyond visual quality metrics, we further evaluate the proposed method on its ability to improve spatially correlated accuracy in image generation. Results on T2I-CompBench indicate that our approach produces more accurate and structurally consistent image content compared to both the base model and the GRPO baseline. As illustrated in Fig.~\ref{fig_example_hps_spatial} and Table~\ref{tab_comp_perference}, the generated images exhibit more precise object boundaries, finer local details, and a clearer understanding of spatial relationships, such as relative positions and object arrangements.

In contrast, although the GRPO baseline improves generation accuracy over the base model to some extent, it often introduces noticeable degradation in visual quality. As shown in Fig.~\ref{fig_example_hps_spatial}, issues such as less coherent room layouts, distorted spatial configurations, and unnatural texture colors can be observed, and performance on certain evaluation categories even slightly decreases. These results suggest that our method better balances spatial reasoning and visual fidelity during RL optimization.

\subsection{Ablations \& Discussions}
We evaluate the effects of proposed loss estimation with multi-diffusion trajectory expectation (``+MTE"), and consistency-aware token selection (``+Sim") on NOVA with the HPS reward, and provide vanilla GRPO (``GRPO") and GRPO with fixing the diffusion decoder (``+Fix Decoder") as comparison, results are as summarized in Table~\ref{tab_ablation}. We analyze the impact of each component below.


\begin{figure*}[t]
\setlength{\abovecaptionskip}{0.1cm}
\setlength{\belowcaptionskip}{0.1cm}
\centering
\begin{minipage}[b]{.31\linewidth}
    \includegraphics[width=1.\linewidth]{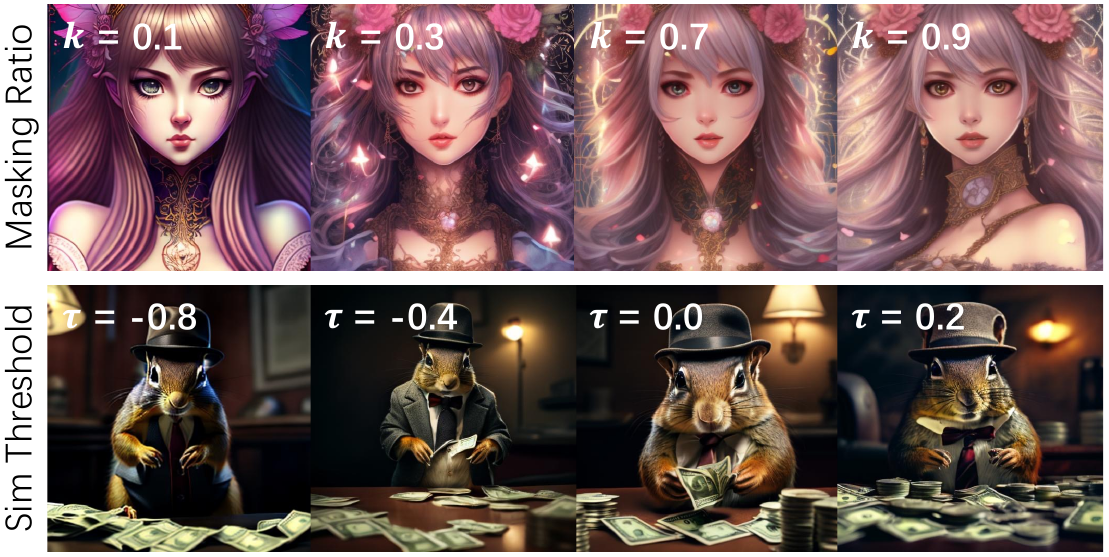}
    \subcaption{Visual of Mask Ratio $\&$ Sim Thresh.}
\end{minipage}\hfill
\begin{minipage}[b]{.35\linewidth}
    \includegraphics[width=1.\linewidth]{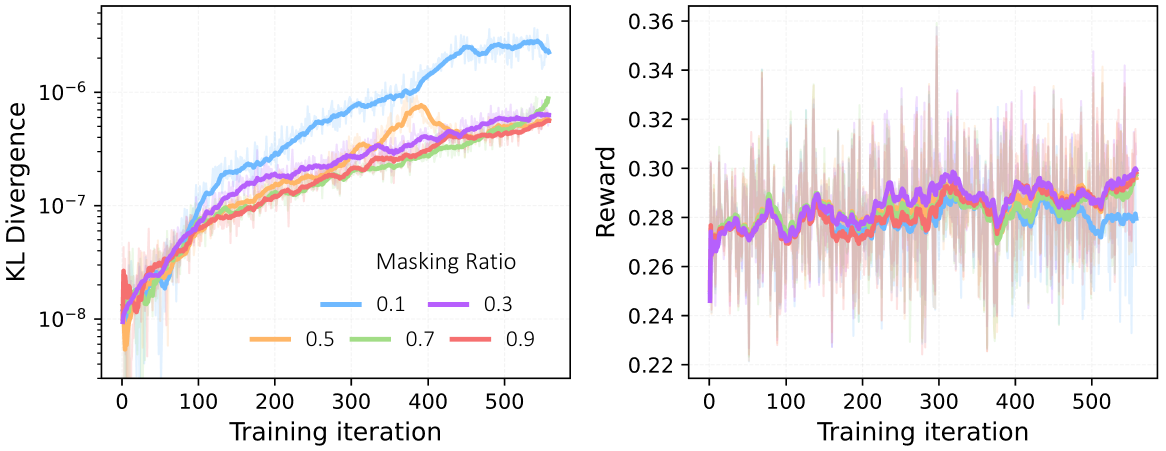}
    \subcaption{Training curves of masking ratios.}
    \label{fig_ab_maskratio_curve}
\end{minipage}\hfill
\begin{minipage}[b]{.33\linewidth}
    \includegraphics[width=1\columnwidth]{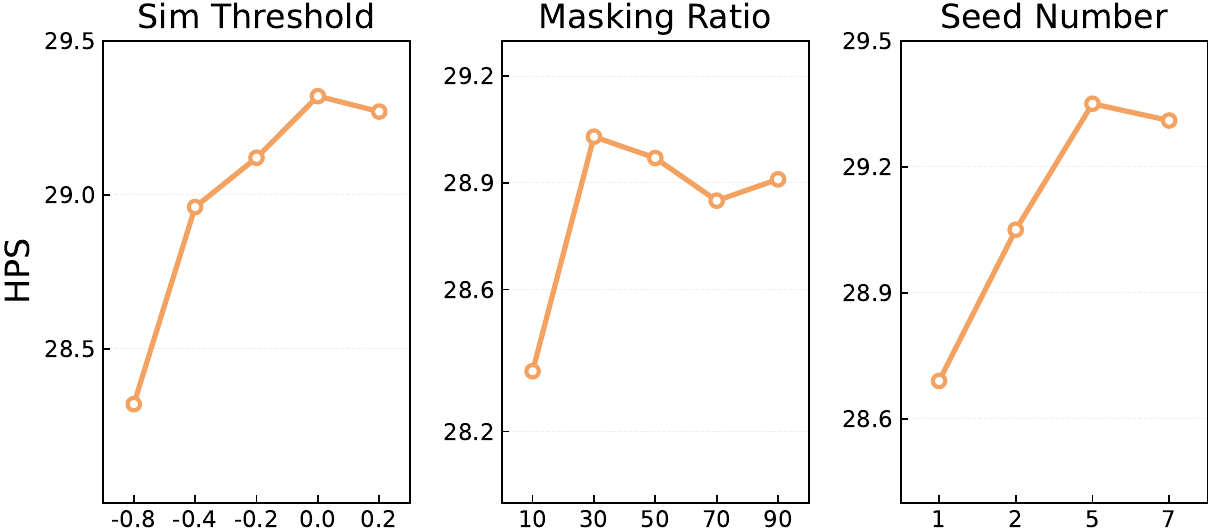}
    \subcaption{Ablations on HPS performance.}
\end{minipage}
\caption{
Ablations of masking ratio, similarity threshold, and diffusion seeds.
(a) Visual results under different masking ratios ($k$) and similarity thresholds ($\tau$), showing trade-off between detail and consistency.
(b) Larger masking ratios improve stability, with marginal gains when $k > 0.3$.
(c) Masking ratio, similarity threshold, and number of diffusion seeds all affect performance.
}
    \vspace{-0.4cm}
    \label{fig_performance_ab}
\end{figure*}
\begin{table}[htbp]
\centering
\setlength{\abovecaptionskip}{0.1cm}
\setlength{\belowcaptionskip}{0.1cm}
\caption{
\textbf{Quantitative results on human preference metrics.}
Compared to vanilla GRPO and its fixed-decoder variant (``+Fix Decoder''), our method, which incorporates multi-trajectory expectation (``+MTE'') and consistency-aware token selection (``+Sim''), achieves improved performance.
Due to the instability of the baseline methods, we report peak performance for ``GRPO'' and ``+Fix Decoder'', where ``*'' indicates the best achieved performance during training.
}
\setlength{\tabcolsep}{2.3mm}{
\resizebox{\linewidth}{!}{
\begin{tabular}{@{}lcc|ccc@{}}
\toprule
            & HPS & Aes. & Shape & Texture & 2d-spat. \\
\midrule
NOVA        & 26.76 & 5.79 & 0.5489 & 0.6712 & 0.2983 \\
~~GRPO*       & 27.57 & 5.78 & 0.5392 & 0.6690 & 0.3069 \\
~~+Fix Decoder*   & 28.65 & 5.83 & 0.5495 & 0.6801 & 0.3134 \\
~~+MTE & 29.06 & 5.90 & 0.5752 & 0.6945 & 0.3289 \\
~~+Sim (Ours) & 29.35 & 6.01 & 0.5780 & 0.6980 & 0.3306 \\
\bottomrule
\end{tabular}
\label{tab_ablation}
}}
\vspace{-10pt}

\end{table}


\noindent\textbf{Loss estimation with multi-trajectory expectation.}
By leveraging expectation of multiple diffusion trajectories with different random seeds, we obtain a more reliable estimate of diffusion log-probabilities, thereby mitigating the adverse impact of diffusion stochasticity on AR policy optimization. This leads to a more stable optimization process and improved policy learning. The quantitative results in Table~\ref{tab_ablation} and Fig.~\ref{fig_example_hps_spatial} further demonstrate the effectiveness and superiority of the proposed approach.

We further analyze training dynamics by visualizing the KL loss and gradient
norms. As illustrated in Fig.~\ref{fig_visualize_ab_seed}, compared to the baseline,
RL with multiple diffusion trajectories yields smaller gradient norms and a more
moderate increase in KL loss in later training stages, indicating more reliable
log-probability estimation. This reduces gradient noise and better preserves the
model distribution during optimization.

\noindent\textbf{Top-$k$\% ratio of selected tokens.}
We further analyze the effect of masking ratio $k\%$ in Sec.~\ref{sec_mte} on training performance. We observe that an excessively high masking ratio leads to over-smoothed training gradients, which slows down RL optimization, whereas an overly low masking ratio results in inaccurate log-probability estimation, causing model to collapse at an early stage of training.
As shown in Fig.~\ref{fig_performance_ab}, evaluating on different masking ratios shows that a $k\!\approx\!30\%$ provides a good balance between training efficiency and stability. In addition, Fig.~\ref{fig_performance_ab}b shows that when more than 30\% of tokens are selected for multi-trajectory estimation, further increasing number of tokens brings no significant performance gain.

\noindent\textbf{Number of trajectories.}
The number of sampled diffusion trajectories also influences RL performance.
Using more diffusion samples provides a more accurate and noise-robust estimate of the diffusion log-probability, thereby improving training stability and final performance. However, excessively many trajectories can over-smooth the optimization gradients.
As shown in Fig.~\ref{fig_performance_ab}c, performance improves as the number of trajectories increases within a moderate range (e.g., < 5 diffusion samples), but saturates and may even degrade when too many trajectories are used, indicating a trade-off between variance reduction and optimization sharpness.

\noindent\textbf{Similarity threshold $\tau$.}
We further study the effect of the threshold $\tau$ used for consistency-aware token selection. Increasing $\tau$ discards more tokens and focuses optimization on fewer ones. As shown in Fig.~\ref{fig_performance_ab}a and Fig.~\ref{fig_performance_ab}c, setting $\tau=0$ yields the best performance, aligns with the intuition that each AR step should progressively move closer to the final image, while overly restrictive token selection may hinder efficiency.

\section{Conclusion}
In this paper, we study reinforcement learning for MAR models and identify the diffusion head as a key bottleneck, as its stochasticity amplifies gradient variance and destabilizes optimization. 
We address this with multi-trajectory expectation (MTE), which reduces diffusion-induced noise, and a consistency-aware token selection strategy that mitigates over-smoothing by focusing on informative tokens. 
Together, they form a noise-corrected GRPO framework that stabilizes training and improves performance. 
Importantly, due to the lightweight nature of the diffusion head, our method introduces negligible additional training overhead. 
Empirically, our approach consistently outperforms strong RL baselines on human preference and spatial-compositional benchmarks.
This work provides an RL framework and analysis for the AR–diffusion hybrid paradigm, paving the way for further research in this direction.

\bibliographystyle{ACM-Reference-Format}
\bibliography{sample-base}

\newpage
\appendix
\onecolumn
\section{Additional Description of Methods}
\subsection{GRPO Optimization for Diffusion Models}
For diffusion models, each denoising step defines a Gaussian posterior transition
$p_\theta(\mathbf{x}_{t-1}\mid \mathbf{x}_t,\hat{\mathbf{x}}_0)
=\mathcal{N}(\boldsymbol{\mu}_\theta,\boldsymbol{\Sigma}_t)$, the posterior is fully determined by the current latent $\mathbf{x}_t$ and the predicted clean latent $\hat{\mathbf{x}}_0$. Specifically, the posterior mean $\boldsymbol{\mu}_\theta(\mathbf{x}_t,\hat{\mathbf{x}}_0,t)$ and variance $\boldsymbol{\Sigma}_t$ are computed from $(\mathbf{x}_t,\hat{\mathbf{x}}_0)$.
The per-timestep log-probability used by our diffusion head is then given by
\begin{equation}
\log p_\theta(\mathbf{x}_{t-1}\mid \mathbf{x}_t, \hat{\mathbf{x}}_0)
=
-\tfrac{1}{2}\!\left(
\frac{\|\mathbf{x}_{t-1}-\boldsymbol{\mu}_\theta\|^2}{\boldsymbol{\Sigma}_t}
+ \log(2\pi \boldsymbol{\Sigma}_t)
\right).
\label{eq_diff_logp}
\end{equation}

We denote $\log p_\theta(\mathbf{x}_{i,t-1}\mid \mathbf{x}_{i,t},\hat{\mathbf{x}}_{i,0})$ as $\log p_{\theta,t}$. Accordingly, the importance ratio is computed as:
\begin{equation}
r_i(\theta)=\exp\!\Big(\sum_t \big(\log p_{\theta,t}-\log p_{\theta_{\mathrm{old}},t}\big)\Big),
\label{eq_diff_importance_ratio}
\end{equation}
For the KL regularization term, we follow the approximation used in prior diffusion-GRPO implementations and employ a Taylor approximation under the fixed-variance Gaussian parameterization above. This gives the following lightweight surrogate:
\begin{equation}
D_{\mathrm{KL}}=\frac{1}{T}\sum_t \Big(
\exp\!\big(\boldsymbol{\mu}_{\mathrm{ref},t}-\boldsymbol{\mu}_{\theta,t}\big)
-\big(\boldsymbol{\mu}_{\mathrm{ref},t}-\boldsymbol{\mu}_{\theta,t}\big)-1
\Big).
\label{eq_diff_kl}
\end{equation}

\subsection{Full Description of Our Proposed Method}
\begin{algorithm*}[h]
\caption{Uncertainty-Aware GRPO for Hybrid AR--Diffusion Models}
\label{alg_uncertainty_grpo_detail}
\begin{algorithmic}[1]
\Require Reference policy $\pi_{\mathrm{ref}}$, prompt dataset $\{c\}$, group size $G$, selected AR mask steps $\mathcal{M}$, selected diffusion steps $\mathcal{T}$, number of diffusion trajectories $S$, uncertainty ratio $k\%$
\Ensure Optimized policy $\pi_\theta$
\State Initialize training policy $\pi_\theta \leftarrow \pi_{\mathrm{ref}}$
\While{not converged}
    \State Set rollout policy $\pi_{\theta_{\mathrm{old}}} \leftarrow \pi_\theta$
    \For{each sampled prompt $c$}
        \State Sample $G$ rollouts $\{o_i\}_{i=1}^G \sim \pi_{\theta_{\mathrm{old}}}(\cdot \mid c)$ and compute group-normalized advantages $\{\hat{A}_i\}_{i=1}^G$
        \For{each sample $i$ and each selected mask step $k \in \mathcal{M}$}
            \State Compute AR latent $Z_i^k$ from previous latent tokens $X_i^{1:k-1}$
            \State Sample $S$ diffusion trajectories $\{X_{i,t}^{k,(s)}\}_{s=1}^S$ conditioned on the same $Z_i^k$
            \State Compute token-wise uncertainty map $u_i^k$ via the standard deviation across $\{X_{i,0}^{k,(s)}\}_{s=1}^S$
            \State Construct binary mask $M_i^k$ by selecting the top-$k\%$ tokens from $u_i^k$
            \For{each selected diffusion step $t \in \mathcal{T}$}
                \State Compute step-wise loss $\ell_{i,k,t}^{(s)}$ for each trajectory $s$ using Eq.~\eqref{eq_important_ratio} and Eq.~\eqref{eq:policy}
                \State Aggregate the uncertainty-aware loss according to Eq.~\eqref{eq_margrpo}
            \EndFor
        \EndFor
        \State Update $\pi_\theta$ using the GRPO objective in Eq.~\eqref{eq_margrpo} over all selected $(i,k,t)$
    \EndFor
\EndWhile
\State \Return $\pi_\theta$
\end{algorithmic}
\end{algorithm*}

\subsection{Additional Visualization Analysis of Latents}
\label{add_vis_latents}
\begin{figure}[h]
\setlength{\abovecaptionskip}{0.1cm}
\setlength{\belowcaptionskip}{0.1cm}
\centering
\resizebox{0.8\linewidth}{!}{%
\begin{minipage}[b]{.38\linewidth}
    \includegraphics[width=\linewidth]{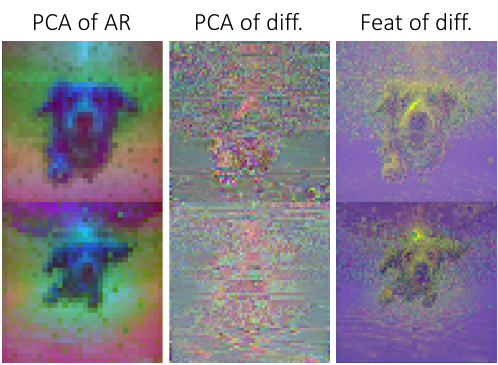}
    \subcaption{Feat. of AR \& diff.}
\end{minipage}\hfill
\begin{minipage}[b]{.59\linewidth}
    \includegraphics[width=1.\linewidth]{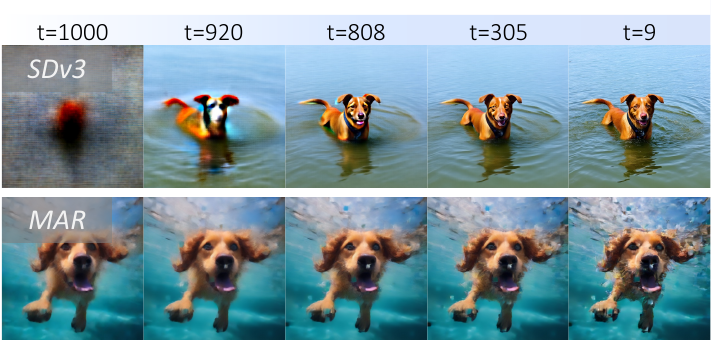}
    \subcaption{Predicted images during sampling.}
\end{minipage}\hfill
}
 \caption{
    (a) AR features encode clear structural information, while diffusion features
    primarily capture fine-grained details. (b) Compared to standard diffusion
    models, the diffusion head in MAR reconstructs most image content at very early
    timesteps.
    } 
    \vspace{-0.3cm}
    \label{fig_motivation_latents}
\end{figure}

\noindent\textbf{Analysis of AR $\&$ diffusion latents.}
As shown in Fig.~\ref{fig_motivation_latents}, AR latents encode clear structural information, whereas
diffusion latents contain more fine-grained details and noise. Moreover, the
diffusion head reconstructs most image content at very early timesteps,
differing from standard diffusion models.

\begin{figure*}[h]
\setlength{\abovecaptionskip}{0.1cm}
\setlength{\belowcaptionskip}{0.1cm}
\centering
\includegraphics[width=0.7\linewidth]{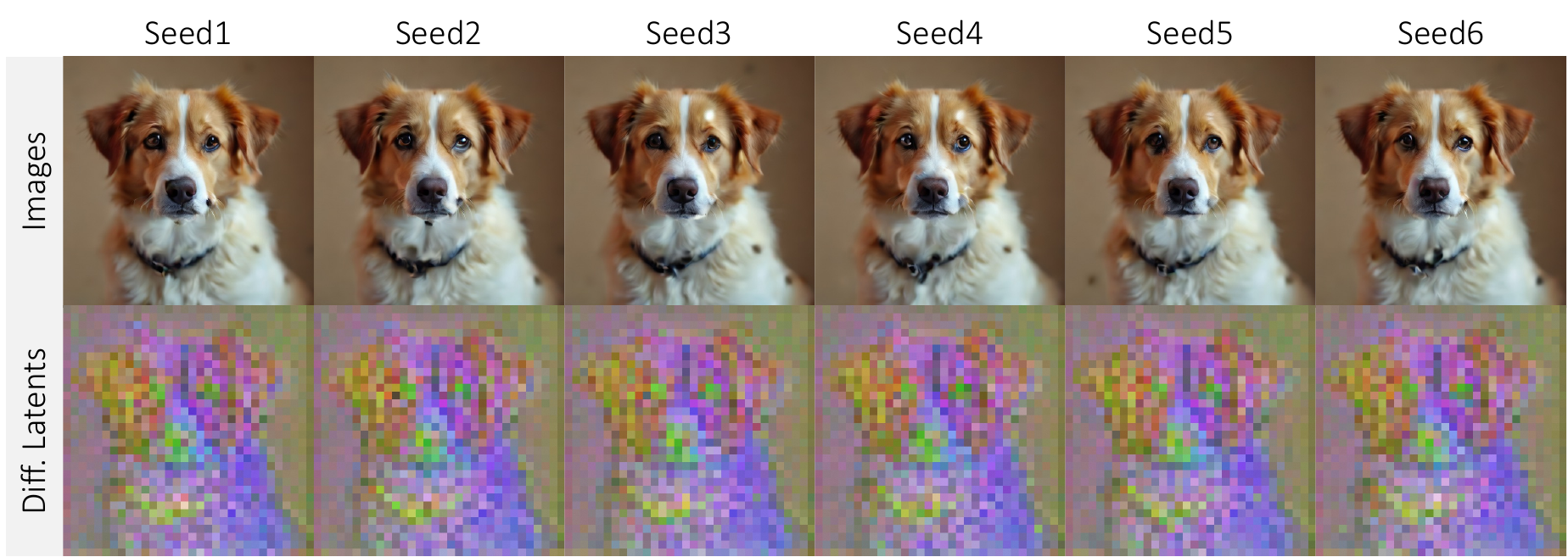}
 \caption{
    Samples generated with different diffusion trajectories is highly deterministic.
    } 
    \vspace{-0.3cm}
    \label{fig_appendix_vis_seed}
\end{figure*}

\noindent\textbf{Samples generated with different diffusion trajectories.}
Conditioned on the same AR latents, varying the random seed during sampling
yields diffusion latents and generated images that are highly consistent, as
illustrated in Fig.~\ref{fig_appendix_vis_seed}.

\section{Detailed Experimental Setup}
\label{appendix_experimental_setup}
We report detailed training configurations, including learning rates and the
number of training steps used for evaluation. For the baseline GRPO, due to its
training instability, we select the checkpoint that achieves the best reward
during training. In contrast, our method is evaluated using a fixed checkpoint
specified in advance.

Additionally, since the diffusion head of Harmon is insufficiently trained, we
perform a brief warm-up stage before RL optimization. Specifically, we fix the
AR component and fine-tune the diffusion head for 400 steps with a learning rate
of $1\times10^{-6}$, followed by the corresponding RL training (for fair comparison, we also conduct a 400 steps finetuning on diffusion head in baseline GRPO). Detailed
numerical settings are provided in Tables~\ref{tab_appendix_setup_nova} and~\ref{tab_appendix_setup_harmon}.

\begin{table}[h]
\centering
\caption{Training hyperparameters for NOVA.}
\setlength{\tabcolsep}{2.5mm}{
\resizebox{0.5\linewidth}{!}{
\begin{tabular}{@{}lcc|cc@{}}
\toprule
               & \multicolumn{2}{c|}{Baseline}               & \multicolumn{2}{c}{Ours} \\ \midrule
Reward func    & HPS                  & Mixed                & HPS         & Mixed      \\ \midrule
Learning-rate  & \multicolumn{2}{c|}{1e-6 (2e-6 collapsed)}  & \multicolumn{2}{c}{2e-6} \\
Training Steps & best ($\sim$250) & best ($\sim$400) & 500     & 800    \\
Beta           & \multicolumn{2}{c|}{0.01}                   & \multicolumn{2}{c}{0.01} \\
Inner loop     & \multicolumn{2}{c|}{1}                      & \multicolumn{2}{c}{1}    \\ \bottomrule
\end{tabular}
\label{tab_appendix_setup_nova}
}}
\end{table}

\begin{table}[h]
\centering
\caption{Training hyperparameters for Harmon.}
\setlength{\tabcolsep}{2.5mm}{
\resizebox{0.5\linewidth}{!}{
\begin{tabular}{@{}lcc|cc@{}}
\toprule
               & \multicolumn{2}{c|}{Baseline}               & \multicolumn{2}{c}{Ours} \\ \midrule
Reward func    & HPS                  & Mixed                & HPS         & Mixed      \\ \midrule
Learning-rate  & \multicolumn{2}{c|}{1e-6}  & \multicolumn{2}{c}{1e-6} \\
Training Steps & best ($\sim$300) & best ($\sim$800) & 500     & 800    \\
Beta           & \multicolumn{2}{c|}{0.01}                   & \multicolumn{2}{c}{0.01} \\
Inner loop     & \multicolumn{2}{c|}{1}                      & \multicolumn{2}{c}{1}    \\ \bottomrule
\end{tabular}
\label{tab_appendix_setup_harmon}
}}
\end{table}

\section{Additional Visualizations \& Analysis}
\subsection{Additional Experiments on Training Dynamics}
In the main text, we demonstrate that fixing the diffusion head yields more stable training than end to end RL optimization. This design is also simpler, as it avoids complex multi stage procedures and delicate learning rate tuning while still providing consistent performance improvements. Here we further investigate different training schemes for model components to better understand why fixing the diffusion head and training only the AR module is necessary for stable GRPO optimization. All experiments are conducted on a simple GRPO baseline to avoid confounding effects from additional techniques such as multi seed strategies.

\begin{figure}[t]
\setlength{\abovecaptionskip}{0.1cm}
\setlength{\belowcaptionskip}{0.1cm}
\begin{center}
\includegraphics[width=0.4\columnwidth]{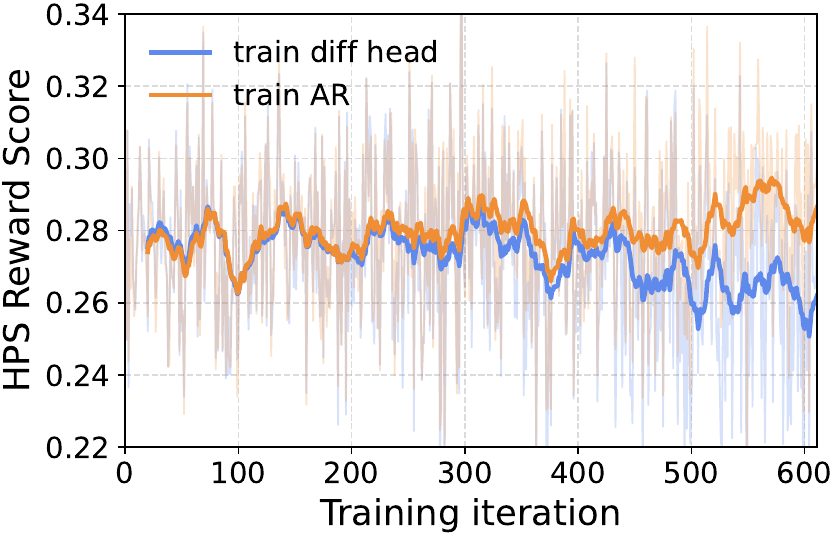}
\end{center}
\caption{
Training dynamics when fixing the AR module and tuning the diffusion head only.
Although the reward can improve slightly at the beginning, it drops quickly afterward,
indicating that optimizing the diffusion head alone is unstable and does not provide
reliable gains for MAR post-training.
}
\label{fig_appendix_train_head_only}
\end{figure}

\noindent\textbf{Training with tuning head only.}
As shown in Fig.~\ref{fig_appendix_train_head_only}, we also test the alternative strategy of fixing the AR module and fine tuning the diffusion head. However, the diffusion head is highly unstable during RL optimization and easily collapses, making it difficult to obtain additional gains through diffusion head tuning alone. 
Even in the Harmon model, we report that the diffusion head must first be fine tuned to avoid poor initial performance in experiments. After reaching a reasonable level, further tuning often leads to degradation, while tuning the AR module instead yields more consistent improvements.

\subsection{Explanation of Why Finetuning on Diffusion Head of Harmon is Needed}
\label{sec_appendix_finetune_harmon}
\begin{figure*}[h]
\setlength{\abovecaptionskip}{0.1cm}
\setlength{\belowcaptionskip}{0.1cm}
\centering
\includegraphics[width=0.8\linewidth]{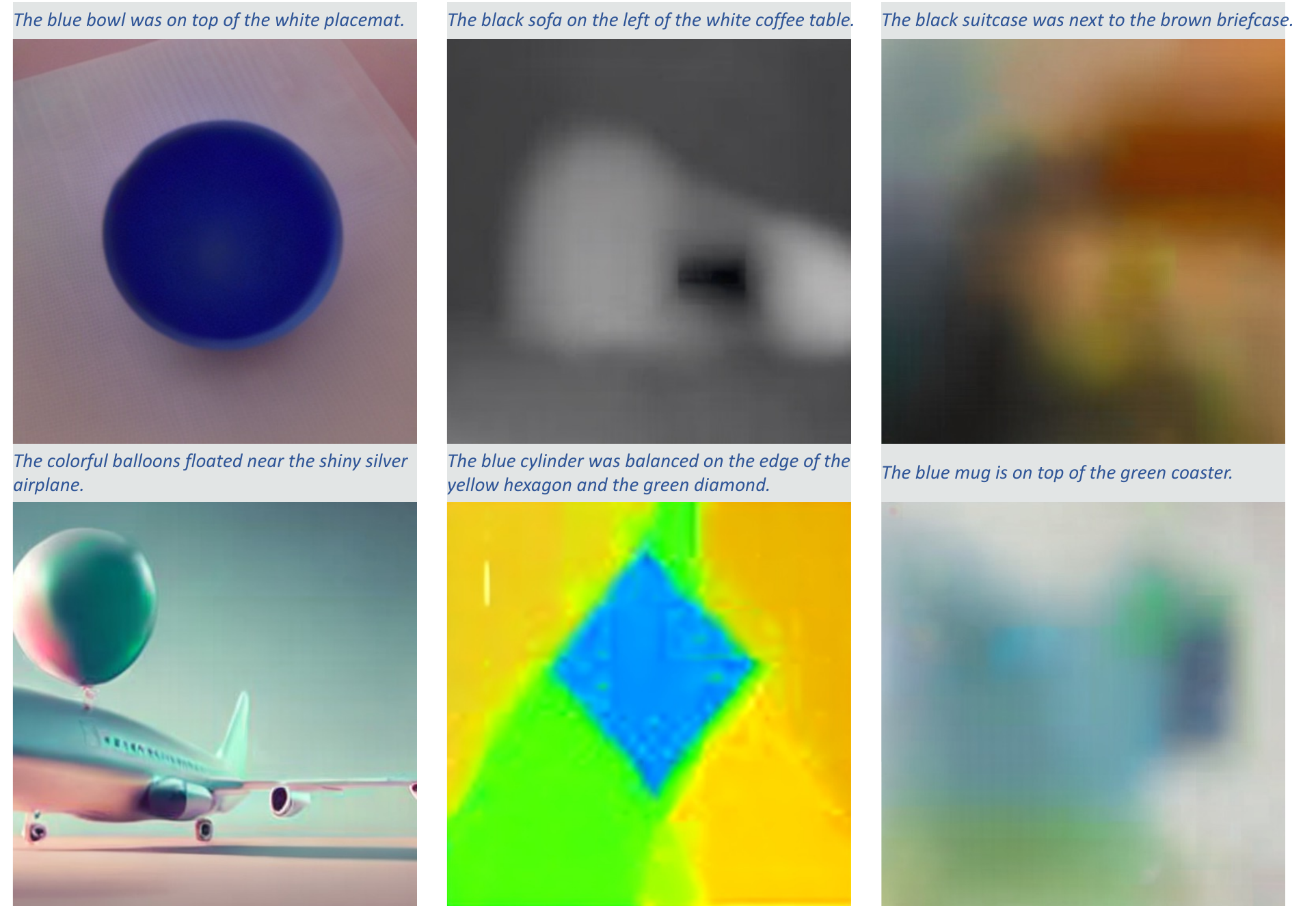}
 \caption{
    Failure cases of the Harmon base model. The model occasionally produces
invalid or severely degraded images due to an insufficiently trained diffusion
head.
    } 
    \vspace{-0.3cm}
    \label{fig_appendix_harmon_badcase}
\end{figure*}

For Harmon, we observe that the base model occasionally fails to produce valid
images in certain cases (see Fig.~\ref{fig_appendix_harmon_badcase}), indicating insufficient training of the
diffusion head. To address this issue, we first apply a short-stage refinement
that updates only the diffusion head (approximately 400 steps), after which the
model is able to generate reasonable images. This suggests that the observed
failure modes primarily stem from an undertrained diffusion head rather than
from the RL algorithm itself. We therefore adopt this lightweight pre-refinement
as a practical initialization before RL, which can be viewed as a form of
task-specific SFT to avoid sub-optimal GRPO caused by a poorly calibrated
diffusion component.

\section{Limitations $\&$ Future Works}
\subsection{Limitations}

Although the proposed method substantially improves the stability and performance ceiling of RL training for MAR models, several limitations remain:

\noindent\textbf{Performance gains vary across different base models.} In our experiments, we observe that the improvement brought by the RL framework differs depending on the underlying model. This is likely due to the discrepancies in diffusion head architectures and pretraining data distributions. In particular, the improvement on Harmon is noticeably smaller than that on NOVA, and the exact cause of this phenomenon remains unclear.

\noindent\textbf{Multiple diffusion trajectories.} The additional diffusion head trajectories are generated by running the diffusion head multiple times with different random seeds after obtaining the full AR latent. This procedure may partially break the causal sampling process, potentially leading to suboptimal RL optimization. Moreover, although the extra forward passes of the diffusion head introduce negligible training time overhead, they may still increase the memory consumption during training.

\subsection{Future Works}
In this work, we investigate the GRPO strategy for MAR models. Building upon our findings, several directions deserve further exploration in future work:

\noindent\textbf{Critic tokens.}
MAR models involve multiple interleaved AR and diffusion steps during generation. 
Consequently, identifying the most informative steps or tokens may be particularly important for effective RL optimization. 
A potential direction is to selectively apply RL signals to a subset of AR or diffusion steps, which may lead to more stable and efficient training. 
Exploring token- or step-level credit assignment in such hybrid generation processes remains an interesting direction for future work.

\noindent\textbf{Video generation.}
Recently, AR--diffusion hybrid frameworks have also been widely adopted in video generation. 
Due to limited computational resources, we do not explore GRPO-based post-training for video models in this work. 
Extending our framework to video generation, where temporal dynamics introduce additional challenges, is a promising direction for future research.

\noindent\textbf{Larger diffusion head.}
Recent unified multimodal generative models tend to employ larger diffusion heads within AR--diffusion hybrid architectures, such as BLIP-3o~\cite{chen2025blip3o}. 
The RL behavior of such models may differ substantially from the lightweight diffusion-head setting studied in this work. 
Investigating RL optimization strategies for models with larger diffusion modules is another important direction for future exploration.

\section{Additional Visualization}
\begin{figure*}[h]
\setlength{\abovecaptionskip}{0.1cm}
\setlength{\belowcaptionskip}{0.1cm}
\centering
\includegraphics[width=1\linewidth]{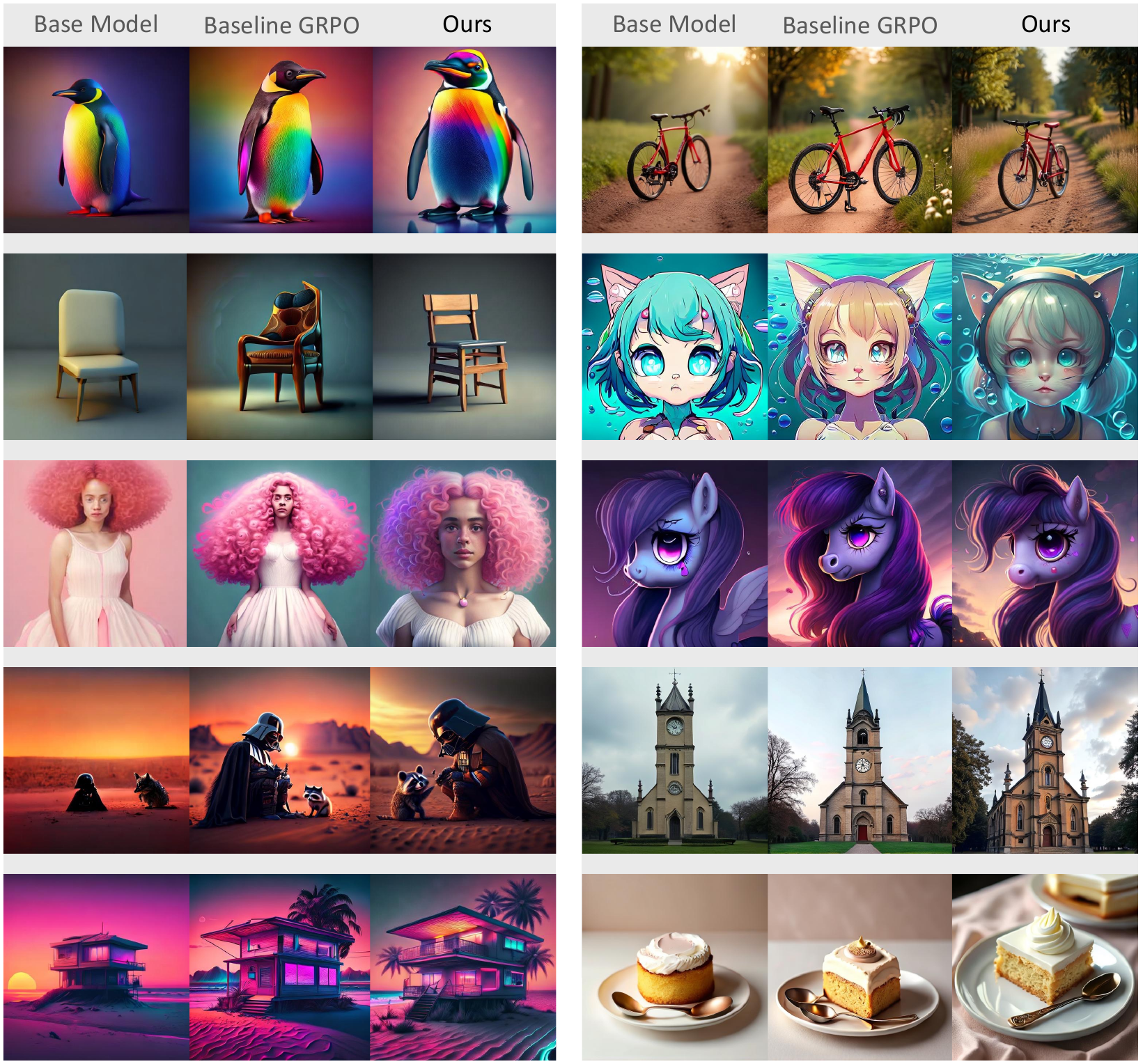}
 \caption{
    Visualization.
    } 
    \vspace{-0.3cm}
    \label{fig_appendix_vis_1}
\end{figure*}

\begin{figure*}[h]
\setlength{\abovecaptionskip}{0.1cm}
\setlength{\belowcaptionskip}{0.1cm}
\centering
\includegraphics[width=1\linewidth]{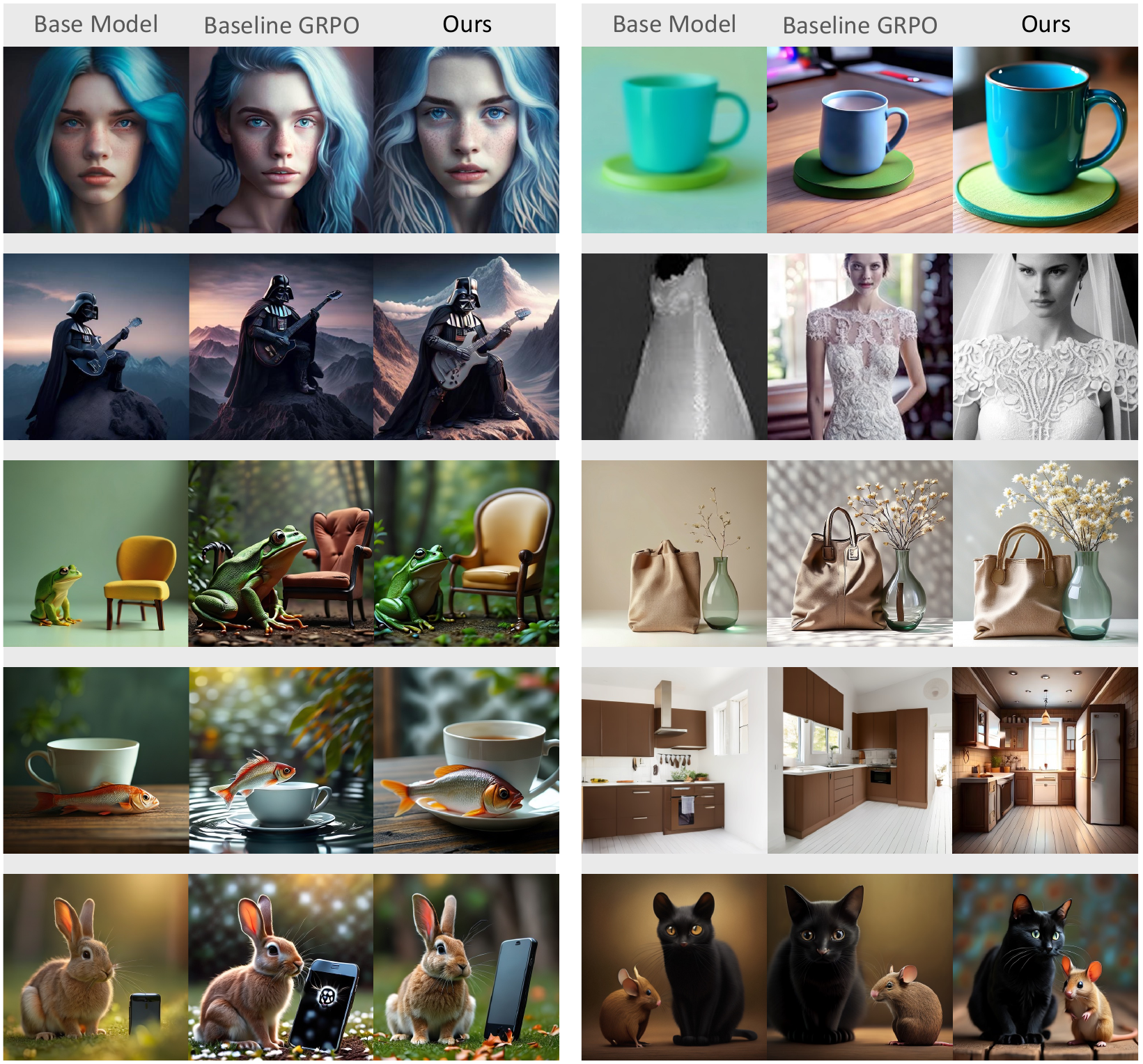}
 \caption{
    Visualization.
    } 
    \vspace{-0.3cm}
    \label{fig_appendix_vis_2}
\end{figure*}

We further present qualitative comparisons of NOVA and Harmon before and after GRPO training to demonstrate the advantage of our method over the Baseline GRPO. As shown in Fig.~\ref{fig_appendix_vis_1}–~\ref{fig_appendix_vis_2}, benefiting from improved optimization stability and better preservation of the model's original distribution, our RL-enhanced results exhibit noticeably richer details, more stable structures, and more coherent content compared to Baseline GRPO.

\end{document}